\documentclass[a4paper,fleqn]{cas-dc}

\usepackage{amsmath,amsfonts,bm}

\def\eqref#1{equation~\ref{#1}}
\def\1{\bm{1}}

\DeclareMathAlphabet{\mathsfit}{\encodingdefault}{\sfdefault}{m}{sl}
\SetMathAlphabet{\mathsfit}{bold}{\encodingdefault}{\sfdefault}{bx}{n}

\newcommand{\E}{\mathbb{E}}

\DeclareMathOperator*{\argmax}{arg\,max}

\def\cH{{\cal H}}
\def\cF{{\cal F}}

\def\cN{{\cal N}}

\def\cX{{\cal X}}

\def\cR{{\cal R}}

\def\cE{{\cal E}}

\def\hat{\widehat}
\newcommand{\mbR}{\mathbb{R}}

\newcommand{\bX}{{\bf X}}
\newcommand{\bx}{{\bf x}}

\newcommand{\bw}{{\bf w}}

\newcommand{\bv}{{\bf v}}

\newcommand{\bfs}{{\bf s}}

\usepackage{lineno,hyperref}
\usepackage{url}
\usepackage{subcaption}
\usepackage{booktabs}
\usepackage{hyperref}
\usepackage{graphicx}
\usepackage{bbm}
\usepackage{epsfig}
\usepackage{fancybox}
\usepackage{pstcol}
\usepackage{multirow}
\usepackage{ulem}
\usepackage{tikz}
\usepackage{wrapfig}
\usepackage{caption}
\usepackage{amsthm}

\newtheorem{theorem}{Theorem}
\newtheorem{lemma}{Lemma}

\newtheorem{definition}{Definition}

\newcommand{\bc}{\begin{center}}
	\newcommand{\ec}{\end{center}}
\newcommand{\be}{\begin{equation}}
	\newcommand{\ee}{\end{equation}}
\newcommand{\ba}{\begin{array}}
	\newcommand{\ea}{\end{array}}
\newcommand{\bean}{\begin{eqnarray*}}
	\newcommand{\eean}{\end{eqnarray*}}
\newcommand{\bea}{\begin{eqnarray}}
	\newcommand{\eea}{\end{eqnarray}}
\newtheorem{Example}{\bf Example}

\newtheorem{proposition}{\bf Proposition}
\newcommand{\ben}{\begin{enumerate}}
	\newcommand{\een}{\end{enumerate}}
\newcommand{\bed}{\begin{itemize}}
	\newcommand{\eed}{\end{itemize}}

\usepackage{amsmath,amsfonts,bm}

\modulolinenumbers[5]
\usepackage[authoryear,longnamesfirst]{natbib}

\def\tsc#1{\csdef{#1}{\textsc{\lowercase{#1}}\xspace}}
\tsc{WGM}
\tsc{QE}
\begin{document}
	\let\WriteBookmarks\relax
	\def\floatpagepagefraction{1}
	\def\textpagefraction{.001}
	
	\shorttitle{SLIDE: a surrogate fairness constraint to ensure fairness consistency}    
	
	\shortauthors{K. Kim, I. Ohn, S. Kim, and Y. Kim}  
	
	\title [mode = title]{SLIDE: a surrogate fairness constraint to ensure fairness consistency}  
	
	\tnotemark[1] 
	
	\tnotetext[1]{This work was supported by Institute for Information \& communications Technology Planning \& Evaluation(IITP) grant funded by the Korea government(MSIT) (No. 2019-0-01396, Development of framework for analyzing, detecting, mitigating of bias in AI model and training data).} 
	
	
	%%%%%%%%%%%%%%%%% first
	
	\author[1]{Kunwoong Kim}[type=editor]
	
	%\cormark[<corr mark no>]
	
	%\fnmark[<footnote mark no>]
	
	\ead{kwkim.online@gmail.com}
	
	% \ead[url]{<URL>}
	
	%\credit{<Credit authorship details>}
	
	\affiliation[1]{organization={Department of Statistics, Seoul National University},
		city={Seoul},
		%          citysep={}, % Uncomment if no comma needed between city and postcode
		postcode={08826}, 
		country={Republic of Korea}}
	
	%%%%%%%%%%%%%%%%% second
	\author[2]{Ilsang Ohn}[type=editor]
	
	% Footnote of the second author
	\fnmark[1]
	
	% Email id of the second author
	\ead{ilsang.ohn@inha.ac.kr}
	
	
	
	\affiliation[2]{organization={Department of Statistics, Inha University},
		city={Incheon},
		%          citysep={}, % Uncomment if no comma needed between city and postcode
		postcode={22212}, % 46556
		%            state={State of Indiana}, % State of Indiana
		country={Republic of Korea}}
	
	\fntext[1]{The work was carried out while being affiliated to the 
		Department of Applied and Computational Mathematics and Statistics, University of Notre Dame, United States of America.}
	
	%%%%%%%%%%%%%%%%%%%% third
	% Footnote of the second author

	\author[3]{Sara Kim}[type=editor]
	\fnmark[2]
	
	% Email id of the second author
	\ead{sarah58833@gmail.com}
	
	
	
	\affiliation[3]{organization={Core Technology R\&D team, Mechatronic R\&D Center, Samsung Electronics},
		city={Hwaseong},
		%          citysep={}, % Uncomment if no comma needed between city and postcode
		postcode={18448}, 
		% state={State of Indiana},
		country={Republic of Korea}}
	
	\fntext[2]{The work was carried out while being affiliated to the 
		Department of Statistics, Seoul National University, Republic of Korea.}

	%%%%%%%%%%%%%%%%%% corr
	\author[1]{Yongdai Kim\corref{corresponding}}[type=editor]
	
	% Footnote of the second author
	
	% Email id of the second author
	\ead{ydkim0903@gmail.com}
	
	
	
	%\affiliation[4]{organization={Department of Statistics, Seoul National University},
	%            % addressline={}, 
	%            city={Seoul},
	%%          citysep={}, % Uncomment if no comma needed between city and postcode
	%            postcode={08826}, 
	%            % state={State of Indiana},
	%            country={Republic of Korea}}

	\cortext[corresponding]{Corresponding author at: Department of Statistics, Seoul National University, Republic of Korea.}
	
	%\fntext[past1]{The work was carried out while being affiliated to the 
	%		Department of Statistics, Seoul National University, Republic of Korea.}
	
	
	\begin{abstract}
		As they have a vital effect on social decision makings, AI algorithms should be not only accurate and but also fair. Among various algorithms for fairness AI, learning a prediction model by minimizing the empirical risk (e.g., cross-entropy) subject to a given fairness constraint has received much attention. To avoid computational difficulty, however, a given fairness constraint is replaced by a surrogate fairness constraint as the 0-1 loss is replaced by a convex surrogate loss for classification problems. In this paper, we investigate the validity of existing surrogate fairness constraints and propose a new surrogate fairness constraint called SLIDE, which is computationally feasible and asymptotically valid in the sense that the learned model satisfies the fairness constraint asymptotically and achieves a fast convergence rate. Numerical experiments confirm that the SLIDE works well for various benchmark datasets.
	\end{abstract}
	
	
	% \begin{highlights}
	% \item 
	% \item 
	% \item 
	% \end{highlights}
	
	% Keywords
	
	\maketitle
	
%	\linenumbers
	
	\section{Introduction}\label{sec1}
	
	Recently, AI (Artificial Intelligence) is being used as decision making tools in various domains such as credit scoring, criminal risk assessment, education of college admissions (\cite{angwin2016machine}) and sentiment analysis (\cite{article-unknow}). As AI has a wide range of influences on human social life, issues of transparency and ethics of AI are emerging.  
	However, it is widely known that due to the existence of historical bias in data against ethics or regulatory frameworks for fairness, trained AI models based on such biased data
	could also impose bias or unfairness against a certain sensitive group (e.g., non-white, women) (\cite{kleinberg2018algorithmic, mehrabi2019survey}). Therefore, designing an AI algorithm that is 
	accurate and fair simultaneously has become a crucial research topic.
	
	The two important issues in fairness AI are definitions of fairness and algorithms to learn fair prediction models.
	There are various definitions of fairness AI, which are roughly categorized into two groups -
	\textit{group fairness} and \textit{individual fairness}. 
	Group fairness (\cite{calders2009building, diref, hardt2016equality}) focuses on treating fairly each sensitive group 
	while individual fairness (\cite{dwork}) focuses on treating fairly any similar individuals. 
	
	Once the definition of fairness is selected, the next step for fairness AI is to choose a learning algorithm to find a fair classifier that is not only accurate but also fair. Most fair learning algorithms belong to one of the three categories -
	pre-processing, in-processing and post-processing methods.
	
	Pre-processing methods remove bias in training dataset or find a fair representation with respect to sensitive variables 
	before the training phase and learn AI models based on de-biased data or fair
	representations.
	\cite{zemel2013learning} proposes a method to learn fair representations based on a clustering algorithm and uses them as feature vectors for solving downstream tasks fairly. After this work, various pre-processing methods including \cite{feldman2015certifying, webster2018mind, xu2018fairgan, creager2019flexibly} have been developed.
	
	In-processing methods generally train an AI model by minimizing the cost function subject to a given fairness constraint.
	Various fairness constraints have been proposed in \cite{kamishima2012fairness, goh2016satisfying, zafar2017fairness, pmlr-v80-agarwal18a, Cotter2019TwoPlayerGF, celis2019classification, zafar2019fairness, cho2020fair, vogel2020learning}, to name just a few.
	In particular, \cite{goh2016satisfying, onconvexand} and \cite{jiang2020wasserstein}  use
	the hinge-surrogate fairness constraint serving as the baseline method in this paper.
	
	Post-processing methods firstly learn an AI model without any fairness constraint and then
	transform the decision boundary or score function of the trained AI model for each sensitive group to satisfy a given fairness constraint.
	For example, \cite{jiang2020wasserstein} proposes to map unfair prediction models to the Wasserstein distance-based barycenter.
	See  \cite{hardt2016equality, kamiran2012decision,fish2016confidence, corbett2017algorithmic, pleiss2017fairness, chzhen2019leveraging, wei20a}
	for other post-processing methods.
	
	In this paper, we are concerned with in-processing fairness AI algorithms.  
	A difficulty in in-processing algorithms is that most fairness constraints involve the indicator function $\mathrm{I}(\cdot>0)$ which
	makes the optimization infeasible.
	To resolve this problem, a popular approach is to replace $\mathrm{I}(\cdot>0)$ by a computationally easier surrogate function such as
	the hinge function $(1+\cdot)_{+}.$ This hinge function is the tight convex upper bound of the indicator function and hence popularly used for a surrogate loss function of the 0-1 loss. Moreover, it is known that the hinge loss is Fisher-consistent in the sense that the minimizer of the population risk with respect to the hinge loss is equal to the Bayes risk (\cite{zhang2004statistical, bartlett2006convexity, blanchard2008statistical}). Thus, we can estimate the Bayes classifier consistently by minimizing the empirical risk with respect to the hinge loss under regularity conditions.
	
	The question we address in this paper is whether this nice property of the hinge function as a surrogate loss of the 0-1 loss is still valid for the fairness constraint. 
	That is, we investigate whether in-processing learning algorithms with a surrogate fairness constraint yield prediction models
	which are (asymptotically) fair in terms of the original fairness constraint.
	Asymptotic properties of fairness AI algorithms have been studied by \cite{pmlr-v65-woodworth17a} and \cite{donini2018empirical}.
	\cite{pmlr-v65-woodworth17a} proposed a two-step procedure to find a prediction model which is
	asymptotically fair. 
	However, as noted by the authors, the proposed algorithm is not computationally feasible since it involves indicator functions in the objective function to be minimized.
	Hence, in practice, the indicator function in constraint should be replaced by a surrogate function.
	
	For representative studies of using surrogate fairness constraint functions,
	the hinge-surrogate (\cite{goh2016satisfying, onconvexand, jiang2020wasserstein}) and
	the linear-surrogate (\cite{donini2018empirical, sensei, chuang2021fair})
	are popularly used. 
	However, no theoretical results about the validity of such surrogate fairness constraints have been studied. 
	Moreover, \cite{pmlr-v119-lohaus20a} raises an issue that using such existing surrogate fairness constraints cannot guarantee a given fairness constraint due to the gap between the original constrained function space and the surrogate one.
	We also provide Figure 2, which illustrates that the hinge-surrogate fairness constraint used in \cite{goh2016satisfying, onconvexand, jiang2020wasserstein} does not provide the optimal model under the original fairness constraint.
	
	We take those fairness surrogate constraints as state-of-the-art baselines in our numerical experiments.
	% We also compare to the linear-surrogate fairness constraint \cite{donini2018empirical, sensei}, which is one of the most common surrogate constraints.
	We refer to the later section of experiments (Section \ref{sec5}) for precise descriptions of baseline methods.
	In summary, a proper surrogate function is highly required for learning fair models.
	
	The aim of this research is to propose a new surrogate fairness constraint that 
	makes the optimization feasible and provides the optimal prediction model under the original fairness constraint asymptotically.
	For this purpose, we develop a new surrogate function called SLIDE for the indicator function.
	We prove that the minimizer of the empirical risk subject to the SLIDE-surrogate fairness constraint 
	is asymptotically equivalent to the minimizer of the population risk under the original
	fairness constraint.   
	
	Our contributions are summarized as follows.
	
	\begin{itemize}
		\item We propose a new surrogate fairness constraint called SLIDE, which is computationally feasible and has desirable theoretical properties.
		
		\item We prove that the SLIDE is an asymptotically valid surrogate fairness constraint by deriving 
		the fairness convergence rate as well as the risk convergence rate of prediction models
		trained by in-processing methods with the SLIDE-surrogate fairness constraint. 
		
		\item We empirically demonstrate by analyzing several benchmark datasets that
		the SLIDE-surrogate fairness constraints are superior to or never worse than the existing surrogate fairness constraints.
		% the hinge-surrogate fairness constraints.
	\end{itemize}
	
	\section{Learning algorithms for fairness AI: Review}\label{sec2}
	
	Let $(Y,\bX,Z)$ be the random vector of a triplet of output, input and sensitive variables,
	whose distribution is $P.$ 
	For simplicity, we consider a binary classification problem (i.e. 
	$Y\in \{-1,1\}$) and a binary sensitive variable (i.e. $Z\in \{0,1\}$). 
	For a given loss function $l$ and a class of prediction models $\cF$, the aim of supervised learning is to find $f^*$ defined as
	$f^*={\rm argmin}_{f\in \cF} \mathbf{E} \left( l(Y,f(\bX)) \right).$
	Due to historical biases or social prejudices, the optimal prediction model $f^*$ would not be socially acceptable because it treats 
	certain groups or individuals unfairly. 
	Thus, we want to search $f$ which is fair 
	and at the same time makes the population risk $\mathbf{E} ( l(Y,f(\bX)) )$ as small as possible.
	
	Suppose that $\cF_{\textup{fair}}$ is a subset of $\cF$ which consists of all fair prediction models. 
	Then, the goal of fair supervised learning 
	is to find $f^*_{\textup{fair}}$ defined as 
	$$ f^*_{\textup{fair}}={\rm argmin}_{f\in \cF_{\textup{fair}}} \mathbf{E} \left( l(Y,f(\bX)) \right). $$
	There exist numerous definitions for the fairness model class $\cF_{\textup{fair}}$, most of which
	can be formulated as
	\be
	\label{eq:constraint}
	\cF_{\textup{fair}}=\{f\in \cF: \phi(f) \le \alpha\}
	\ee
	for a positive constant $\alpha,$ where
	$\phi: \cF \rightarrow [0,\infty)$ is a so called \textit{fairness constraint function} corresponding to a given definition of fairness.
	In the next subsection, two representative fairness constraints
	are explained, which we focus for theoretical derivations and empirical studies.
	Our theoretical results, however, can be applied to most of other fairness constraints without much modification.
	
	\subsection{Examples of fairness constraints}
	
	We consider two representative fairness constraints - one for group fairness and the other for individual fairness. For a given real-valued function $f$ and an input $\mathbf{X},$ the corresponding classifier is constructed by ${\rm sign} \{f(\mathbf{X})\}.$
	
	{\bf Disparate Impact:} The first fairness constraint is disparate impact (DI) (\cite{diref}). 
	A prediction model $f$ satisfies the {\it $\alpha$-DI} if
	$\phi(f)\le \alpha,$ where
	\be
	\label{eq:di}
	\phi(f) = | \mathbf{P} \left( f(\bX) \ge 0 | Z=0 \right) - \mathbf{P} \left( f(\bX) \ge 0 | Z=1 \right) |.
	\ee
	Other group fairness constraints can be defined by replacing $\mathbf{P}( f(\bX) \ge 0| Z=z)$ in (\ref{eq:di}) for $z \in \{0,1\} $ by
	other conditional probabilities. For example, the equalized odds (EO) and equal opportunity (EqOpp) from \cite{hardt2016equality} are defined by replacing $ \mathbf{P} ( f(\bX) \ge 0| Z=z)$
	into $ \mathbf{P} ( f(\bX) \ge 0| Z=z, Y=y)$ for $y\in \{-1,1\} $ and $\mathbf{P} ( f(\bX) \ge 0| Z=z, Y=1)$ respectively. 
	
	{\bf Uniform Individual Fairness:} Individual fairness requires that
	similar individuals should be treated similarly. \cite{dwork} introduces the initial notion of individual fairness: we say a classifier $f \in \mathcal{F}$ is individually fair if it satisfies the Lipschitz property, i.e., for every $ \mathbf{x}, \mathbf{x}' \in \mathcal{X}$, $ D(f(\mathbf{x}), f(\mathbf{x}')) \le d(\mathbf{x}, \mathbf{x}') $ with respect to a similarity metric $D(\cdot, \cdot)$  between prediction models and a similarity metric 
	$d(\cdot, \cdot)$ between individuals. \cite{pacf} introduces a relaxed notion: a classifier $f \in \mathcal{F}$ is $(\alpha, \gamma)$-approximately individually fair if it satisfies $ \mathbf{P}_{\mathbf{x}, \mathbf{x}'} ( D(f(\mathbf{x}), f(\mathbf{x}')) - d(\mathbf{x}, \mathbf{x}')> \gamma ) \le \alpha$ for a given $\gamma>0.$ 
	In this paper, we consider a new definition of individual fairness called 
	{\it  $(\alpha,\gamma,\epsilon)$-uniform individual fair (UIF)} defined as
	$\phi(f ; \gamma,\epsilon) \le \alpha,$ where
	\begin{equation}
		\label{eq:uif}
		\phi(f ; \gamma,\epsilon) := \mathbf{P} \big( \sup_{ \mathbf{v} : d(\mathbf{X}, \mathbf{v}) \le \epsilon} D \left( f(\mathbf{X}), f(\mathbf{v}) \right) > \gamma \big).
	\end{equation}
	The definition of UIF is motivated by the definition of SenSeI from \cite{sensei} which requires that
	$$ \mathbf{E} \big( \sup_{\bv:d(\bX,\bv) \le \epsilon} D(f(\bX),f(\bv)) \big) \le \alpha.$$
	We replace the expectation by the probability even though one more parameter (i.e. $\gamma$) is needed.
	In Section \ref{sec5}, we show that UIF is better than SenSeI.
	
	\subsection{Learning algorithms for fairness AI}\label{sec2-2}
	
	Let $(y_1,\bx_1,z_1),\ldots, (y_n,\bx_n,z_n)$ be given training dataset which are assumed to be
	independent realizations of $(Y,\bX,Z).$ Let $\phi_n$ be the empirical version of $\phi.$ 
	The $\phi_n$ for DI and UIF are given as
	\be
	\label{eq:di-emp}
	\phi_n(f)=\left| \frac{1}{n_0} \sum_{i: z_i=0} \mathrm{I}(f(\bx_i)>0) -\frac{1}{n_1} \sum_{i: z_i=1} \mathrm{I}(f(\bx_i)>0)\right|
	\ee
	and
	\begin{equation}
		\label{eq:uif-emp}
		\phi_n(f ; \gamma,\epsilon) = \frac{1}{n} \sum_{i=1}^n \mathrm{I} \big( D \left( f(\bx_i), f(\bv_i') \right) > \gamma \big)
	\end{equation}
	respectively,
	where $n_z=\sum_{i=1}^n \mathrm{I}(z_i=z)$ and 
	$\bv_i'=\argmax_{\bv: d(\bx_i,\bv) \le \epsilon} D(f(\bx_i),f(\bv)).$
	
	Let $\cF_{n,\alpha}=\{f\in \cF:\phi_n(f)\le \alpha\}.$
	Most in-processing fair learning algorithms try to  minimize the empirical risk
	$L_n(f)$ on $\cF_{n,\alpha},$ where $L_n(f)=\sum_{i=1}^n l(y_i,f(\bx_i))/n.$
	However, this optimization is hard since $\phi_n$ is not continuous. Typically this problem is resolved by use of a surrogate fairness constraint. One of the most popular surrogate fairness constraint is to replace
	the indicator function $\mathrm{I}(\cdot\ge 0)$ in $\phi_n$
	by the hinge function $(1+\cdot)_{+}$, which we denote $\phi_n^{\textup{hinge}}(f)$ and $\phi_n^{\textup{hinge}}(f ; \gamma,\epsilon).$
	Then, we learn a prediction model by minimizing the empirical risk subject to
	$\phi_n^{\textup{hinge}}(f)\le \alpha$ which is similar to \cite{goh2016satisfying, onconvexand} and \cite{jiang2020wasserstein}. 
	We note that not only for the hinge-surrogate constraint, but one can also consider other existing surrogate constraints that are used in \cite{zafar2017fairness, donini2018empirical, Madras2018LearningAF} and \cite{chuang2021fair}.
	
	\section{SLIDE: A new surrogate fairness constraint}
	\label{sec3}
	
	For a fixed $\alpha > 0$ and any $\delta>0,$ we say that a trained prediction model $\hat{f}_n$ is fairness-consistent if
	$ \mathbf{P} \{ \phi(\hat{f}_n)\le \alpha+\delta \} \rightarrow 1$
	as $n\rightarrow \infty.$
	The aim of this section is to propose a new surrogate fairness constraint with which
	the corresponding (in-processing) trained prediction model is fairness-consistent.
	
	The hinge function is popularly used as a surrogate loss function of the 0-1 loss 
	in classification problems, and it is shown that the resulting estimator is risk-consistent in the sense that the mis-classification error
	of the trained prediction models converges to that of the Bayes classifier  (\cite{zhang2004statistical,bartlett2006convexity,blanchard2008statistical}). 
	This nice property of the hinge function, however, would not hold for fairness-consistency. 
	This is mainly because the surrogate fairness constraint may not be asymptotically equivalent to the original fairness constraint. 
	To resolve this problem, we propose a new surrogate function so-called SLIDE for the indicator function $\mathrm{I}(\cdot>0)$ such that the corresponding surrogate fairness constraint is asymptotically equivalent to the original fairness constraint, and thus the resulting prediction model becomes fairness-consistent as well as risk-consistent.

	\subsection{Proposed surrogate fairness constraint: SLIDE}
	\label{sec4-1}
	
\begin{figure}[h]
		\centering
		\includegraphics[width=0.3\textwidth]{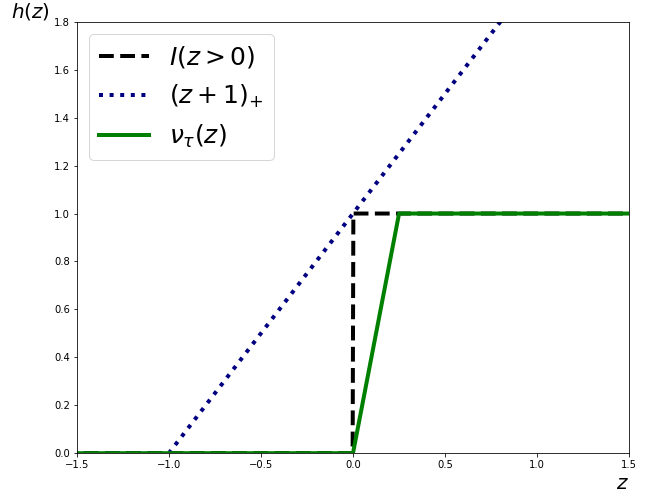}
		\caption{Comparison of the 0-1, hinge and SLIDE ($\tau = 0.25$) functions (black long dotted line: the $0$-$1$, blue short dotted line: the hinge, and green solid line: the SLIDE).}
		\label{fig1}
	\end{figure}

	For a given $\tau>0,$ the SLIDE function $\nu_\tau(\cdot): \mathbb{R}\rightarrow [0,1]$ is defined as
	\be
	\label{eq:slide}
	\nu_\tau(z)=\frac{z}{\tau} \mathrm{I} (0< z \le \tau) + \mathrm{I} (z>\tau).
	\ee
	Figure \ref{fig1} compares the $0$-$1$, the hinge and SLIDE functions. 
	The function $\nu_{\tau}$ looks similar to a slide so that we call it SLIDE. 
	Note that the SLIDE function is a lower bound of the $0$-$1$ function while the hinge function is an upper bound. In addition, the SLIDE function is non-convex while the hinge function is convex. The non-convexity would make the corresponding optimization more difficult, but our experiments suggest that standard gradient descent based optimization algorithms work well with the SLIDE function. 
	Moreover, it is possible to apply convex-concave procedure (CCCP) of \cite{cccp} since $\nu_\tau(\cdot)$ is decomposed by the sum of convex and concave functions. 
	See sections of experiments (Section \ref{sec5} and \ref{sec:E}) for details.
	The empirical SLIDE-surrogate fairness constraints for DI and UIF are obtained by replacing $\mathrm{I}(\cdot>0)$ in $\phi_n$ by $\nu_\tau(\cdot),$
	which we denote $\phi_{n,\tau}^{\textup{slide}}(f)$ and $\phi_{n,\tau}^{\textup{slide}}(f ; \gamma,\epsilon),$ respectively.
	We also denote $\phi^{\textup{slide}}_{\tau}(f)$ and $\phi^{\textup{slide}}_{\tau}(f ; \gamma, \epsilon)$ as population version of both, respectively.
	For notational simplicity, we omit the superscript ``slide'' when the meaning is clear.
	
	The SLIDE function is motivated by the $\Psi$ learning of \cite{shen2003psi}, where $\Psi(z)= ( z / \tau )  \cdot \mathrm{I} (-\tau< z \le 0)  + \mathrm{I} (z > 0)$
	is used as a surrogate loss of the negative 0-1 loss. Even though the $\Psi$ function is an upper bound of the indicator function,
	it would not be appropriate for a surrogate fairness constraint since $\phi_n(f)$ depends on samples with $-\tau< f(\bx_i)<0$ for some $i \in \{ 1, \cdots, n\}.$
	We modify the $\Psi$ function to have the SLIDE function.

	\subsection{Comparison of the SLIDE- and hinge-surrogate fairness constraints with a toy example}
	
	Let $\cF_{n,\alpha}^{\textup{hinge}}=\{f\in \cF: \phi_n^{\textup{hinge}}(f)\le \alpha\}$ and $\cF_{n,\alpha,\tau}^{\textup{slide}}=\{f\in \cF: \phi_{n,\tau}^{\textup{slide}}(f) \le \alpha\}.$
	In this section, by analyzing a toy example, we illustrate that $\cF_{n,\alpha,\tau}^{\textup{slide}}$ is closer to the original constraint class $\cF_{n,\alpha}$ than $\cF_{n,\alpha}^{\textup{hinge}}$ is.
	
\begin{figure}[t]
		\centering
		\includegraphics[scale = 0.155]{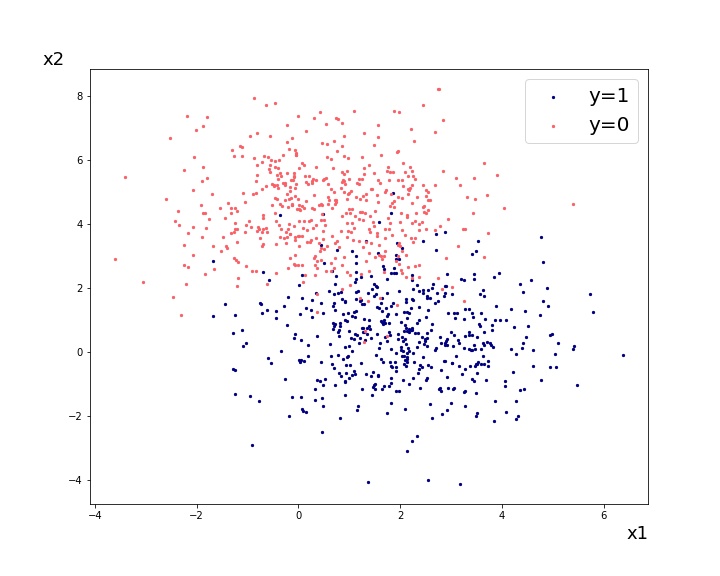}
		\includegraphics[scale = 0.235]{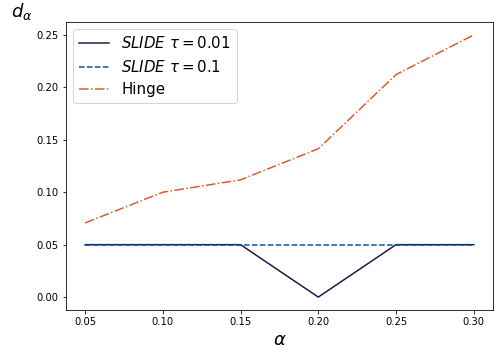}
		\caption{(Left) The scatter plot of the 2-D toy dataset (2-component Gaussian Mixture). (Right) Plot of $\alpha$ vs. $d^{\textup{hinge}}_\alpha$ and $d^{\textup{slide}}_{\alpha,\tau}$ for $\tau\in \{0.01, 0.1\}$.}
		\label{fig2}
	\end{figure}

	The left panel of Figure \ref{fig2} presents the toy dataset with two dimensional samples
	which are generated from a mixture of two Gaussian distributions whose details are given in Appendix.
	%This panel is given to show how the distribution of samples look like.
	We consider linear model as $\cF=\{f_\beta (\bx)=\beta_1 x_1 + \beta_2 x_2:  \beta:=(\beta_1,\beta_2)^\top\in (-2,2)^2\}.$
	For the fairness constraint, we consider the relaxed individual fairness in \cite{pacf} as
	$\phi_{\textup{if}}(f ; \gamma)= \mathbf{P}_{\bX,\bX'} \{ D(f(\bX),f(\bX'))-d(\bX,\bX')>\gamma \}$
	with $\gamma=0.3,$
	where $\bX' $ is a independent copy of $\bX.$
	
	We calculate $\cF_{n,\alpha},\cF_{n,\alpha}^{\textup{hinge}}$ and $\cF_{n,\alpha,\tau}^{\textup{slide}}$
	by a grid search and compare them as follows. 
	Let $\Theta_{n,\alpha}=\{\beta: f_\beta \in \cF_{n,\alpha}\},$ and
	$\Theta_{n,\alpha}^{\textup{hinge}}$ and $\Theta_{n,\alpha,\tau}^{\textup{slide}}$ are defined accordingly.
	We obtain those sets of fair parameters by calculating the population version of the fairness constraint values (i.e., $\phi_{\textup{if}}(f; \gamma)$ ) based on Monte-Carlo simulation at the selected parameters on grids (i.e., $(\beta_{1}, \beta_{2})$ on the $200 \times 200$ grids of $(-2, 2) \times (-2, 2)).$
	%That is, we directly check whether a given $(\beta_{1}, \beta_{2})$ satisfies the three fairness constraints.
	
	Let $D_H(\cdot,\cdot)$ be the Hausdorff distance between two subsets in $(-2, 2)^2.$
	It turns out that 
	$$ D_{H}(\Theta, \Theta') = \max \left( D_{H, 1}, D_{H, 2} \right)
	$$
	where $D_{H, 1} := \sup_{ \theta \in \Theta } \inf_{ \theta' \in \Theta' } || \theta - \theta' ||_{2}$ and $D_{H, 2} := \sup_{ \theta' \in \Theta' } \inf_{ \theta \in \Theta } || \theta' - \theta ||_{2}$
	for two subsets $\Theta$ and $\Theta'.$
	We calculate $d^{\textup{hinge}}_\alpha=\min_{\alpha'} D_{H}(\Theta_{n,\alpha}, 
	\Theta_{n,\alpha'}^{\textup{hinge}})$ and $d^{\textup{slide}}_{\alpha,\tau}=\min_{\alpha'} D_{H}(\Theta_{n,\alpha}, \Theta_{n,\alpha',\tau}^{\textup{slide}})$ for each $\alpha.$
	Then we draw the plot of $\alpha$ versus 
	$d^{\textup{hinge}}_\alpha$ and $d^{\textup{slide}}_{\alpha,\tau}$ with $\tau\in \{0.01, 0.1\}$ which is given in the right panel of Figure \ref{fig2}. 
	We consider $\alpha$ less than 0.3 since the level of fairness for the optimal classifier is around 0.3.
	Note that $d^{\textup{hinge}}_\alpha$ is getting larger as $\alpha$ increases
	while $d^{\textup{slide}}_{\alpha,\tau}$ stays at a lower level regardless of $\alpha.$ 
	That is, the hinge-surrogate fairness constraint does not approximate the original fairness constraint well 
	%when $\alpha$ is close to 0.3 
	and thus the corresponding fair prediction model would be suboptimal.
	In contrast, the SLIDE-surrogate fairness constraint approximates the original fairness constraint relatively well. This phenomenon is still observed for the two-moon dataset and for DI, whose results are given in Appendix \ref{sec:E}.
	
	\section{Theoretical analysis}
	\label{sec4}
	
	In this section, we consider the estimated prediction model $\hat{f}_n$ obtained by minimizing the empirical risk $L_n(f)$ on
	$\cF_{n,\alpha+\delta_n, \tau_{n}}^{\textup{slide}}=\{ f\in \cF: \phi_{n,\tau_n}^{\textup{slide}}(f)\le \alpha+\delta_n\}$
	for given $\tau_n$ and  $\delta_n$ converging to 0,
	and study asymptotic properties of $\hat{f}_n.$
	In particular, we derive the upper bound of $\tau_n$ such that the fair prediction model with the SLIDE-surrogate fairness constraint is asymptotically equivalent to the fair prediction model with the original fairness constraint. 
	Note that a larger $\tau_n$ is better in view of computation
	since the SLIDE function becomes less non-convex. 
	For the sake of readers' convenient, we list the mathematical notations and symbols introduced in the previous sections in Table \ref{tablea1}.
	
	\begin{table*}[h]
		\footnotesize
		\centering
		\caption{A list of mathematical notations and symbols.}
		\begin{center}
			\begin{tabular}{c|c}
				\toprule
				Notation (Reference if exists) & Description \\
				\midrule
				\midrule
				$ \mathbf{X} \in \mathbb{R}^{d} $ & Input random vector \\
				$ Y \in \{-1, 1\} $ & Output random variable \\
				$ Z \in \{0, 1\} $ & Sensitive random variable \\
				\midrule
				$f$ & A prediction model \\
				$\hat{f}_{n}$ & The fair model learned by the SLIDE-surrogate fairness constraint \\
				$f_{\alpha}^{\star}$ & The true fair model \\
				\midrule
				$ \phi $ (Equations (\ref{eq:di}), (\ref{eq:uif})) & A fairness constraint function \\
				$ \phi_{\tau}^{\textup{slide}} $ & The SLIDE-surrogate constraint function for a given $\phi$ \\
				$ \phi^{\textup{hinge}} $ & The hinge-surrogate constraint function for a given $\phi$ \\
				% \midrule
				$ \alpha $ (Equation (\ref{eq:constraint})) & Upper bound of constraint function, i.e., level of fairness \\
				$ \gamma $ (Equation (\ref{eq:uif})) & The relaxation parameter for UIF \\
				$ \epsilon $ (Equation (\ref{eq:uif})) & The maximum perturbation norm for UIF \\
				\midrule
				$ \nu_{\tau} $ (Equation (\ref{eq:slide})) & The SLIDE function with $\tau > 0$ \\
				\bottomrule
			\end{tabular}
		\end{center}
		\label{tablea1}
	\end{table*}

	For a given estimator $\hat{f}_n,$
	we say that the fairness convergence rate of $\hat{f}_n$ is $a_n$ if
	$\phi(\hat{f}_n)\le \alpha+a_n$ in probability.
	Similarly, we say that the {$l$-excess} risk convergence rate of $\hat{f}_n$ is $b_n$ if
	$\cE_{l} (\hat{f}_n,f^\star_\alpha) := \mathbf{E} ( l(Y, \hat{f}_n(\bX)) ) -\mathbf{E} ( l(Y,f^\star_\alpha(\bX)) ) \le b_n$
	in probability, where $f^\star_\alpha$ is the true minimizer of $\mathbf{E} ( l(Y,f(\bX)) )$
	among all measurable functions $f$ with $\phi(f)\le \alpha.$
	We derive $a_n$ and $b_n$ in terms of $\tau_n$ and $\delta_n$ as well as the complexity of $\cF,$ that provide a guide to choose $\tau_n$ and $\delta_n.$ 
	We allow the class of models $\cF$ to depend on the sample size, denoted by $\cF_{n},$ 
	which is popularly used in nonparametric regression contexts to avoid overfitting. 
	For notational simplicity, we drop the subscript $n$ whenever the meaning is clear.

	For technical reasons, we derive the fairness and $l-$excess convergence rates of $\hat{f}_n$
	which depend on $\hat{f}_n$ itself. To be more specific, we assume that there exists a constant $M_f > 0$ 
	depending on $f\in \cF$	such that
	$$|\phi(f)-\phi_{\tau_n}^{\textup{slide}}(f)| \le M_f \tau_n.$$
	%	For example, we can let $M_{f} =  \sup_{r \in [0, \tau_{n}], z \in \{ 0, 1 \}} g_{f|z} (r)$ for DI,
	%   where $g_{f|z}$ is the density of $f(\mathbf{X})$ conditional on $Z = z.$
	For example of UIF, we can set $M_{f} =  \sup_{r \in [-\nu,\nu]} g_{D} (r),$
	where $g_{D}$ is the density of $ \sup_{ \mathbf{v} : d(\mathbf{X}, \mathbf{v}) \le \epsilon } D( f(\mathbf{X}), f(\mathbf{v}) )$ and $\nu$ is a positive constant greater
	than $\tau_n.$ Our convergence rates depend on $M_{\hat{f}_n}$ as well as $\tau_n.$

	\subsection{Fairness convergence rate}
	
	The fairness convergence rate depends on the two quantities: (1) the complexity of $\cF$ and (2) the choice of $\tau_n$ and $\delta_n.$ 
	For the complexity of $\cF,$ the empirical Rademacher complexity defined as
	$$
	\hat{\cR}(\cF)=\frac{1}{n} \mathbf{E}_{\mathbf{\sigma}_{i} \sim U (\{\pm1\})^n} \left( \sup_{f\in\cF}\sum_{i=1}^n\sigma_if(\bX_i) \right)
	$$ is a standard measure. 
	The following two theorems show that how the fairness convergence rates depend on
	$\tau_n$ and $\delta_n$ as well as the empirical Rademacher complexity.	
	
	\begin{theorem}[Fairness convergence rate for DI]
		\label{thm:fair_conv_di}
		%Let $\hat{f}_n$ be the minimizer of the empirical risk $L_n(f)$ on$\cF_{n,\alpha+\delta_n}^{\textup{slide}}=\{ f\in \cF: \phi_{n,\tau_n}(f)\le \alpha+\delta_n\}.$ Then
		When $\phi$ is DI, the fairness convergence rate of $\hat{f}_n$ is given by
		\begin{equation*}
			a_n= \mathcal{O} \left(\delta_n+ M_{\hat{f}_{n}}\tau_n+\sum_{z\in\{0,1\}}\hat\cR_z(\nu_{\tau_n}(\cF))+\sqrt{\frac{\log n}{n}}\right),
		\end{equation*}
		where $\nu_{\tau_n}(\cF):=\{\nu_{\tau_n}(f):f\in\cF\}.$
	\end{theorem}
	
	%Let
	%\begin{equation*}
	%    \psi_{\gamma,\epsilon}(f)=\sup_{\bx': d(\bX,\bx')\le\epsilon} D(f(\bX),f(\bx'))-\gamma.
	%\end{equation*}

	\begin{theorem}[Fairness convergence rate for UIF]
		\label{thm:fair_conv_uif}
		When $\phi$ is UIF with the parameters $\gamma$ and $\epsilon,$ the fairness convergence rate of  
		$\hat{f}_{n}$ is given by
		\begin{equation*}
			a_n= \mathcal{O} \left(\delta_n+ M_{\hat{f}_{n}}\tau_n+\hat\cR(\nu_{\tau_n}\circ \eta(\cF))+\sqrt{\frac{\log n}{n}}\right),
		\end{equation*}
		where $\nu_{\tau_n}\circ \eta(\cF)=\{\nu_{\tau_n}\circ\eta_f:f\in\cF\}$ and
		$\eta_f(\bx) := D(f(\bx),f(\bx'))-\gamma.$
	\end{theorem}
	
	Note that the fairness convergence rates in Theorems \ref{thm:fair_conv_di} and 
	\ref{thm:fair_conv_uif} cannot be faster by $\sqrt{\log n/n}.$
	Theorems \ref{thm:fair_conv_di} and  \ref{thm:fair_conv_uif} imply that the largest $\tau_n$ without sacrificing the fairness convergence
	rate is $\mathcal{O}(\sqrt{\log n/n}).$ A situation is similar for $\delta_n$ in the sense
	that $\delta_n= \mathcal{O}(\sqrt{\log n/n})$ makes the constrained function space 
	$\cF_{n, \alpha+\delta_n, \tau_{n}}^{\textup{slide}}$ be as large as possible without affecting the fairness convergence rate.
	If the Rademacher complexities are larger than $\mathcal{O}(\sqrt{\log n/n}),$
	we can let $\tau_n$ and $\delta_n$ be even larger.
	
	Desirable upper bounds of the Rademacher complexities could not be derived directly from that of $\cF$ since the Lipschitz constant of the SLIDE function $\nu_{\tau_n}$ diverges as $n\rightarrow\infty.$
	We calculate the upper bounds of the Rademacher complexities by calculating the metric entropy of
	$\nu_\tau \circ \eta(\cF)$ directly. 
	The results for the case of $\cF$ being linear model class are provided in Appendix.
	For deep neural networks, we consider a case where the class $\cF$ of prediction models
	depends on the sample size $n$ and calculate empirical Rademacher complexities
	to derive ($l$-excess) risk convergence rates in the following subsection.
	
	\subsection{Risk convergence rate}

	The convergence rate of the ($l$-excess) risk depends on various choices including the loss function
	and the model class $\cF.$ 
	We use the logistic loss $l(y,f)=\log(1+\exp(-yf)).$ For $\cF,$  we consider deep neural networks with the ReLU activation, $L_n$ many layers, $N_n$ many nodes at each layer, $S_n$ many non-zero weights and biases that are bounded by $B_n$, the final output value being bounded by $F_n.$ 
	To derive the risk convergence rate, we assume that $f^\star_\alpha$ belongs to the H\"{o}lder space with smoothness $\zeta$ (see the definition of H\"{o}lder space in Appendix). The next theorem derives the risk convergence rates for DI and UIF as well as the fairness convergence rates. 
	
	\begin{theorem}[Risk convergence rates for DI and UIF]
		\label{thm:risk_conv_di}
		Suppose that $n_1/n_0\to s\in(0,\infty)$ and that $\phi(f)$ is $M$-Lipschitz with respect to $\|\cdot\|_\infty.$ That is,
		$|\phi(f_1)-\phi(f_2)|\le M \|f_1 -f_2\|_\infty$ for some constant $M > 0.$ Let $b_n:=n^{-\frac{\zeta}{2\zeta + d}} (\log n)^{3/2}$. Moreover, assume that $\delta_n/b_n\rightarrow \infty$ as $n\rightarrow\infty.$ Then, for both DI and UIF, there exist positive sequences $L_n, N_n,S_n,B_n$ and $F_n$ such that
		\begin{align*}
			\cE(\hat{f}_n, f^\star_\alpha)&\le \mathcal{O}(b_{n} + M_{\hat{f}_{n}} \tau_n)\\
			\phi(\hat{f}_n) &\le \alpha + \mathcal{O}(\delta_{n} + b_{n} + M_{\hat{f}_{n}} \tau_n)
		\end{align*}
		with probability at least $1-1/n.$
	\end{theorem}
	
	Assume that $M_{\hat{f}_n}$ is bounded.	
	The largest rate of $\tau_n$ that minimizes the risk convergence rate and fairness convergence rate simultaneously is
	$b_n$ (i.e., $\tau_{n} = \mathcal{O}(b_{n})$), which makes the two convergence rates be almost equal provided $\delta_n/b_n$ diverges slowly (e.g., $\log n$ order).
	This result implies that we can set $\tau_n$ larger when $\cF$ is more complex (i.e., $\zeta$ is smaller and hence $b_n$ is larger)
	and vice versa.
	
	For standard classification problems, the risk convergence rate
	could be faster than $1/\sqrt{n}$ and hence the risk convergence rate in Theorem \ref{thm:risk_conv_di}
	looks suboptimal. However, this slower convergence rate is unavoidable since $f^\star_\alpha$ is not the global minimizer of
	the population risk $\mathbf{E} \{ l(Y,f(\bX)) \}$ among all measurable functions. We believe that the risk convergence rate
	in Theorem \ref{thm:risk_conv_di} would be optimal.
	Here, we note that Table \ref{tablea1} in Appendix summarizes mathematical notations used in this section.
	
	\subsection{Remarks on $M_{\hat{f}_n}$}\label{Mf}
	
	The convergence rates in Theorems \ref{thm:fair_conv_di}, \ref{thm:fair_conv_uif} and \ref{thm:risk_conv_di} would be meaningful
	only if $M_{\hat{f}_n}$ is not too large (i.e., bounded).
	Let $\tau_n = \mathcal{O}(b_n),$ which is the largest $\tau_n$ that does not change the convergence rates. 
	In Proposition 1 of Appendix, we prove that
	$M_{\hat{f}_n} \le M_{n,\hat{f}_n}+\mathcal{O} (1),$ where
	$$
	M_{n,f}=| \phi_{n, -\tau_{n}}^{\textup{slide}} (f; \gamma, \epsilon) - \phi_{n, \tau_{n}}^{\textup{slide}}(f; \gamma, \epsilon) |/\tau_n
	$$ for UIF, where $\phi_{n, -\tau_{n}}^{\textup{slide}}$ is the opposite SLIDE-surrogate fairness constraint (i.e., $\nu_{\tau}$ in $\phi_{n, \tau_{n}}^{\textup{slide}}$ is replaced by $\nu_{-\tau} = \frac{z}{\tau} \mathrm{I} ( - \tau < z \le 0) + \mathrm{I} (z>0)$). The population version of $\phi_{-\tau_{n}}^{\textup{slide}}$ can be defined similarly.
	A formula of $M_{n,f}$ for DI is provided in Proposition \ref{prop2} of Appendix. Thus, when $M_{n, \hat{f}_n}$ is not large, we expect that $M_{\hat{f}_n}$ is also small.
	
	The above result provides a way of using the SLIDE-surrogate fairness constraint in practice.
	First, we learn $f$ by $\hat{f}_n$ with the SLIDE-surrogate fairness constraint. Then, if $M_{n,\hat{f}_n}$
	is not too large, we keep using $\hat{f}_n$ for prediction. Otherwise, we abort $\hat{f}_n$ and resort to other fairness AI algorithms. For example, we decide that $M_{n,\hat{f}_n}$ is too large if $M_{n,\hat{f}_n} \tau_n$ is larger than 10\% of $\phi(\hat{f}_n).$ 
	In Table \ref{table8} in Appendix E,
	we observe that $M_{n,\hat{f}_n} \tau_n$ is not larger than 10\% of $\phi(\hat{f}_n)$ for the datasets analyzed in our numerical studies,
	which amply supports the validity of the SLIDE-surrogate fairness constraints.

	\section{Experiments}
	\label{sec5}

	We compare the SLIDE-surrogate fairness constraint to the hinge-surrogate fairness constraint
	by analyzing three benchmark datasets for both group fairness and individual fairness. 
	For the class of models, we use single hidden layer deep neural networks. 
	%We utilize the \texttt{Pytorch} framework with NVIDIA TITAN RTX GPUs. 
	Details about learning algorithms are described in Appendix.

	\paragraph{Datasets}
	\label{para:data}
	We analyze three public datasets popularly used in fairness AI: (1) \textit{Adult} dataset from \cite{adultdata}, (2) \textit{Bank} dataset from \cite{uci}, and (3) \textit{Law} dataset from SEAPHE \footnote{http://www.seaphe.org/databases.php}. They have gender, age and race as the sensitive variable, respectively. 
	For each dataset, we split the whole dataset into training, validation and test datasets with ratio 60\% : 20\% : 20\% randomly and repeat this random splitting three times.

	\paragraph{Performance measures}		
	For prediction accuracy, we use two measures: accuracy (Acc) and balanced accuracy (BA). 
	The balanced accuracy, which is considered by \cite{Yurochkin2020Training, sensei}, is an average of the accuracies in each class of label $y =-1$ and $y = 1.$ For assessing fairness, we calculate DI on test data for group fairness. For individual fairness, we use the consistency (Con.) of prediction considered by \cite{Yurochkin2020Training, sensei}. 
	Con. measures how frequently the predictive class labels of two inputs that are the same except the sensitive variable coincide. Note that a larger value of Con. means that the prediction model is more individually fair.
	
	\paragraph{Optimization algorithms}
	
	To learn a fair prediction model, we minimize $L_n(f)+\lambda \phi_{n}(f)$ on $f\in \cF$ 
	for a given surrogate fairness constraint $\phi_n,$ where $\lambda$ is the Lagrangian multiplier.
	For $\lambda,$ we fix the accuracy at a certain level and choose $\lambda$ such that
	such that the accuracy of $\hat{f}_{n,\lambda},$ a (local) minimizer of $L_n(f)+\lambda \phi_{n}(f),$ 
	on the validation data is closest to the fixed accuracy. Then, we compare the level of fairness.
	Since deep neural networks and the SLIDE function are highly non-linear,
	we train multiple models with multiple random initial parameters
	and then select the most fair model. We use the Adam optimizer of \cite{adam} for the optimization.
	
	For the SLIDE-surrogate, we try another optimization algorithm where the solutions obtained with the hinge-surrogate fairness constraint are used as initial solutions, which we call the hybrid SLIDE (HySLIDE) algorithm.
	By comparing the SLIDE (a gradient descent algorithm with random initials) and the HySLIDE, we can investigate how sensitive the SLIDE-surrogate fairness constraint is to the choice of initials. 
	In addition, we apply the CCCP algorithm of \cite{cccp} to the SLIDE + UIF, whose details and results are provided in Appendix.
	
	\begin{figure}[h]
		\scriptsize
		\centering
		\includegraphics[scale = 0.27]{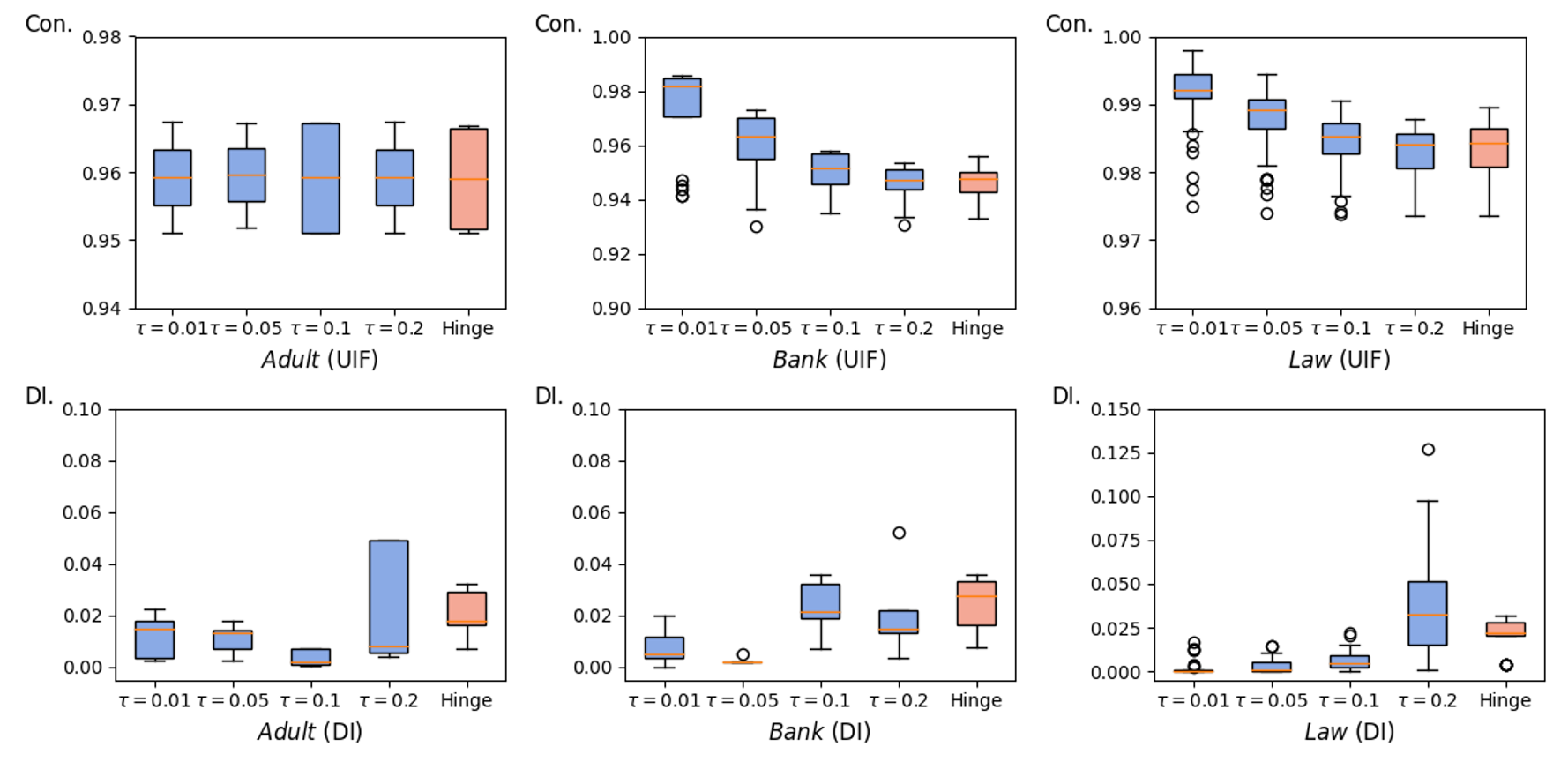}
		\caption{ Box plots of levels of fairness (upper - Con., lower - DI) of fair models learned by the SLIDE-surrogate constraint with $\tau \in \{ 0.01, 0.05, 0.1, 0.2 \}$ and the hinge-surrogate constraint on \textit{Adult}, \textit{Bank}, and \textit{Law} test datasets. }\label{fig:3}
	\end{figure}
	
	Figure \ref{fig:3} draws the box plots of the levels of fairness
	of the fair prediction models learned with the SLIDE-surrogate and hinge-surrogate fairness constraints.
	The implication of Figure \ref{fig:3} is that performance of the SLIDE does not strongly depend on the value of $\tau$ as long as $\tau$ lies in a reasonable range (i.e., $(0.01, 0.2)$). 
	
	The SLIDE-surrogate fairness constraint has one more regularization parameter $\tau$ compared to the hinge-surrogate fairness constraint. To make the comparison fair, we select $\tau$ randomly from $(0.01, 0.2)$ along with a random initial solution. For the hinge-surrogate fairness constraint, we learn multiple prediction models corresponding to multiple random initial parameters and choose the most fair model among those whose accuracies on the validation dataset are similar to a priori given accuracy.
	Similarly, for the SLIDE-surrogate fairness constraint, we learn multiple prediction models corresponding to multiple pairs of randomly selected $\tau$ and randomly selected initial parameters and choose the most fair model.
	
	% We choose the regularization parameters on validation data and assess the performances (accuracy and level of fairness) on test data.
	
	%That is, we set the number of multiple prediction models to be the same as $10$ for SLIDE and other fairness constraints to make the comparison be fair.	

	%\paragraph{Choice of  $\tau_n$}
	%Motivated by the results of Figure 3, we also choose $\tau_n$ randomly from the range between 0.01 and 0.2 for each random initial even there exists some variations of performances with respect to the choice of $\tau_{n}.$
	
	\begin{table}[h]
		\caption[9pt]{Group fairness (DI) performances (Acc(\%), BA(\%), DI) of fair models learned by the SLIDE-surrogate and the hinge-surrogate constraints on \textit{Adult}, \textit{Bank} and \textit{Law} test datasets.}
		\label{table1}
		\footnotesize
		\centering
		\begin{center}
			\begin{tabular}{c|c|cc|c|}
				\toprule
				Dataset & Method & Acc & BA & DI \\
				\midrule
				\midrule
				\multirow{3}{*}{\textit{Adult}} & DI + Hinge & 82.80 & 72.34 & .013 (.001) \\
				& DI + SLIDE & 82.19 & 72.14 & \textbf{.007} (.003) \\
				& DI + HySLIDE & 82.24 & 72.33 & .010 (.002) \\
				\midrule
				\multirow{3}{*}{\textit{Bank}} & DI + Hinge & 90.07 & 73.33 & .015 (.002) \\
				& DI + SLIDE & 90.24 & 73.22 & \textbf{.013} (.002) \\
				& DI + HySLIDE & 90.01 & 73.42 & .016 (.003) \\
				\midrule
				\multirow{3}{*}{\textit{Law}} & DI + Hinge & 82.51 & 59.24 & .013 (.004) \\
				& DI + SLIDE & 82.49 & 58.99 & \textbf{.008} (.003) \\
				& DI + HySLIDE & 82.41 & 59.54 & .014 (.003) \\
				\bottomrule
			\end{tabular}
		\end{center}
	\end{table}	
	
	\begin{table}[h]
		\caption[9pt]{Group fairness (DI) performances (Acc(\%), BA(\%), DI) of fair models learned by the state-of-the-art baseline methods (DI + Cov from \cite{zafar2017fairness} and DI + Linear from \cite{donini2018empirical}) on \textit{Adult}, \textit{Bank} and \textit{Law} test datasets.}
		\label{table2}
		\footnotesize
		\centering
		\begin{center}
			\begin{tabular}{c|c|cc|c|}
				\toprule
				Dataset & Method & Acc & BA & DI \\
				\midrule
				\midrule
				\multirow{3}{*}{\textit{Adult}} & DI + Cov  & 82.11 & 72.50 & .018 (.003) \\
				& DI + Linear  & 82.09 & 72.05 & .016 (.003) \\
				& DI + SLIDE & 82.19 & 72.14 & \textbf{.007} (.003) \\
				\midrule
				\multirow{3}{*}{\textit{Bank}} & DI + Cov  & 90.04 & 73.41 & .029 (.004) \\
				& DI + Linear  & 89.35 & 73.35 & .016 (.002) \\
				& DI + SLIDE & 90.24 & 73.22 & \textbf{.013} (.002) \\
				\midrule
				\multirow{3}{*}{\textit{Law}} & DI + Cov  & 82.50 & 58.95 & .020 (.003) \\
				& DI + Linear  & 82.40 & 58.83 & .014 (.001) \\
				& DI + SLIDE & 82.49 & 58.99 & \textbf{.008} (.003) \\
				\bottomrule
			\end{tabular}
		\end{center}
	\end{table}
	
	\begin{figure}[h]
		\scriptsize
		\centering
		\includegraphics[scale = 0.255]{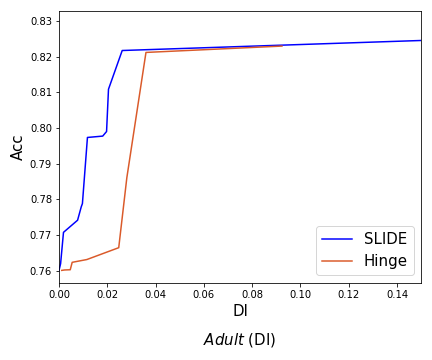}
		\includegraphics[scale = 0.255]{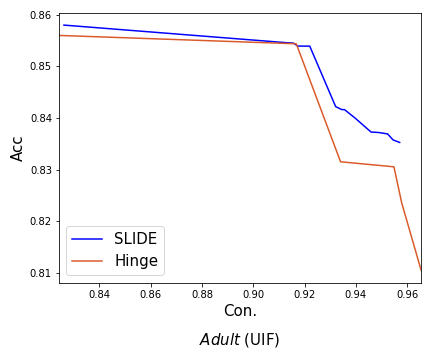}
		\\
		\includegraphics[scale = 0.255]{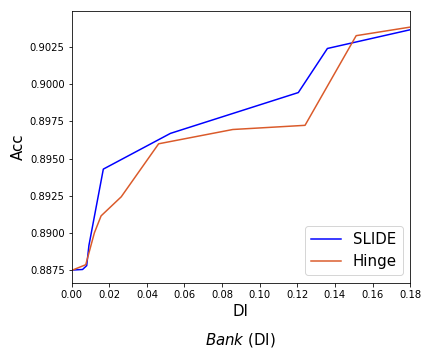}
		\includegraphics[scale = 0.255]{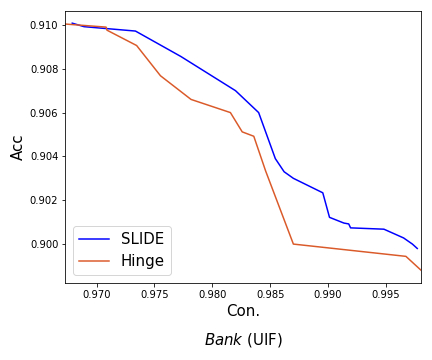}
		\\
		\includegraphics[scale = 0.255]{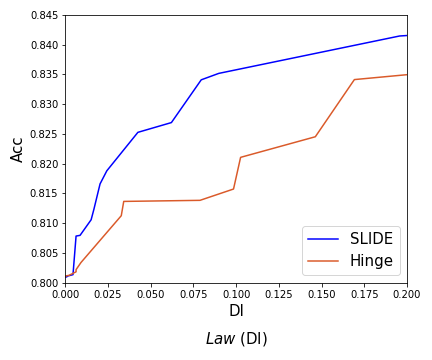}
		\includegraphics[scale = 0.255]{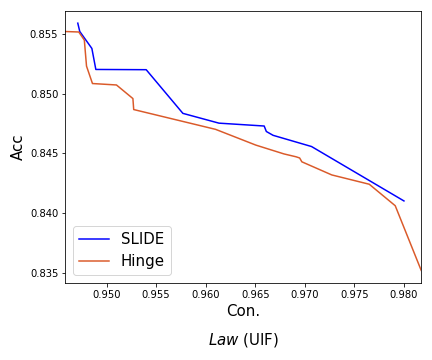}
		\caption{ \textbf{(Left)} Group fairness (DI) Pareto-front lines between DI and Acc on (upper): \textit{Adult}, (center): \textit{Bank}, and (lower): \textit{Law} test datasets. 
			\textbf{(Right)} Individual fairness (UIF) Pareto-front lines between Con. and Acc on (upper): \textit{Adult}, (center): \textit{Bank}, and (lower): \textit{Law} test datasets.
			These results are averaged on each hyperparameter $\lambda.$
			The blue lines are the results of learned models by the SLIDE-surrogate constraint, and the orange lines are those by the hinge-surrogate constraint.
		}
		\label{fig4}
	\end{figure}
	
	\subsection{Group fairness}
	
	We compare the SLIDE-surrogate and hinge-surrogate fairness constraints for DI.
	Table \ref{table1} summarizes the results, which indicate that the SLIDE-surrogate fairness constraint performs better than the hinge-surrogate fairness constraint for DI.
	In addition, it is noticeable that the SLIDE outperforms the HySLIDE.
	
	Furthermore, we compare the SLIDE-surrogate fairness constraint with other state-of-the-art methods, \cite{zafar2017fairness} and \cite{donini2018empirical} in Table \ref{table2}.
	\cite{zafar2017fairness} used a covariance-surrogate constraint for DI (DI + Cov), 
	and \cite{donini2018empirical} used a linear-surrogate constraint for equalized odds so that we simply modified it to DI (DI + Linear).
	%so that we abbreviately denote those methods as DI + Cov and DI + Linear, respectively in Table 2.
	In Table \ref{table2}, the SLIDE-surrogate fairness constraint outperforms the other state-of-the-art competitors.
	
	For comparing the overall performances of the DI + SLIDE and the DI + Hinge,
	we draw the Pareto-front lines between the DI values and accuracies 
	corresponding to various values of $\lambda$
	in the left side of Figure \ref{fig4}. 
	The lines are the averages taken over five repetitions of learning with five random initials.
	It is obvious that the SLIDE-surrogate uniformly dominates the hinge-surrogate in that the SLIDE-surrogate provides a better trade-offs between DI and Acc.
	
	%The superiority of SLIDE-surrogate fairness constraint is still observed if we change the accuracies of the prediction models 
	%(recall that we compare the levels of fairness while the accuracies are set to be similar)
	%to be lower or higher than those used in our experiments, whose results are presented in Appendix.
	
	\subsection{Individual fairness}
	
	We compare the SLIDE-surrogate and hinge-surrogate fairness constraints for UIF.
	Table \ref{table3} compares the SLIDE-surrogate and the hinge-surrogate fairness constraints
	as well as the HySLIDE, which indicates that the SLIDE is superior to the hinge for UIF
	and competitive to the HySLIDE. Recall that the HySLIDE is inferior to the SLIDE in Table \ref{table1}. The results in Tables \ref{table1} and \ref{table3} together 
	imply at least that the SLIDE is not sensitive to the choice of the initial parameters.
	
\begin{figure}[h]
		\scriptsize
		\centering
		\includegraphics[scale = 0.27]{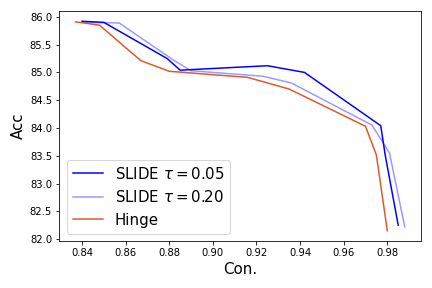}
		\includegraphics[scale = 0.27]{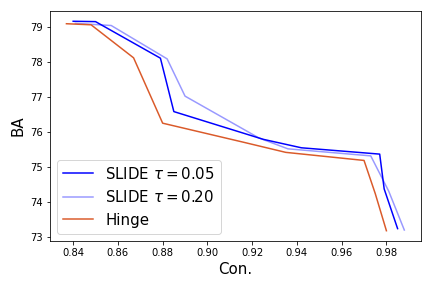}
		\caption{Individual fairness (UIF) Pareto-front lines between (left) Con. and Acc, and (right) Con. and BA on \textit{Adult} test dataset.
			Blue and skyblue lines are those for the SLIDE-surrogate with different $\tau = 0.05, 0.20,$
			and the orange lines are those for the hinge-surrogate.}
		\label{fig5}
	\end{figure}

	Figure \ref{fig5} draws the Pareto-front lines between the accuracy or the balanced accuracy and a level of fairness, i.e., the consistency (Con.) for the \textit{Adult} dataset, 
	which clearly shows that the SLIDE-surrogate fairness constraint uniformly dominates the hinge-surrogate fairness constraint in that provides a better trade-off. 
	
	Table \ref{table4} compares  two other individual fairness learning algorithms - SenSR (\cite{Yurochkin2020Training}) and SenSeI (\cite{sensei}), i.e., UIF + Linear.
	% with the surrogate UIF constraints.
	Here, we follow the data pre-processing and regularization parameter selection technique used in  \cite{Yurochkin2020Training} and \cite{sensei}. 
	In Table \ref{table4}, S-Con. is the consistency when the variable ``spouse'' is changed and
	GR-con. is the consistency when the variables ``gender'' and ``race'' are changed (\cite{Yurochkin2020Training, sensei}). 
	For these two fairness measures, the SLIDE-surrogate fairness constraint performs better than SenSR and UIF + Linear as well as the hinge-surrogate fairness constraint. 
	
	For comparing mean performances on regularization parameter, we also run five experiments with five different initial parameters, and take averages on each $\lambda.$
	Then, we compute the Pareto-front lines which is displayed in the right side of Figure \ref{fig4}.
	It is also observed that the SLIDE-surrogate outperforms the hinge-surrogate in that the SLIDE-surrogate provides a better trade-offs between Con. and Acc.
	
	\begin{table}[h]
		\caption[9pt]{Individual fairness (UIF) performances (Acc(\%), BA(\%), Con.) of fair models learned by the SLIDE-surrogate and the hinge-surrogate constraints on \textit{Adult}, \textit{Bank} and \textit{Law} test datasets.}
		\label{table3}
		\footnotesize
		\centering
		\begin{center}
			\begin{tabular}{c|c|cc|c|}
				\toprule
				Dataset & Method & Acc & BA & Con. \\
				\midrule
				\midrule
				\multirow{3}{*}{\textit{Adult}} & UIF + Hinge & 84.60 & 75.75 & .916 (.003) \\
				& UIF + SLIDE & 84.36 & 75.73 & .920 (.005) \\
				& UIF + HySLIDE & 84.51 & 75.69 & \textbf{.922} (.003) \\
				\midrule
				\multirow{3}{*}{\textit{Bank}} & UIF + Hinge & 90.29 & 63.82 & .985 (.006) \\
				& UIF + SLIDE & 90.14 & 63.43 & \textbf{.990} (.005) \\
				& UIF + HySLIDE & 90.15 & 63.03 & .984 (.004) \\
				\midrule
				\multirow{3}{*}{\textit{Law}} & UIF + Hinge & 83.75 & 62.67 & .985 (.006) \\
				& UIF + SLIDE & 83.99 & 62.71 & .986 (.004) \\
				& UIF + HySLIDE & 84.02 & 62.79 & \textbf{.987} (.003) \\
				\bottomrule
			\end{tabular}
		\end{center}
	\end{table}
	
	\begin{table}[h]
		\caption[9pt]{Individual fairness (UIF) performances (Acc(\%), BA(\%), S-Con., and GR-Con.) of fair models learned by the state-of-the-art baseline methods (UIF + Linear from \cite{sensei} and SenSR from \cite{Yurochkin2020Training}) on \textit{Adult} test dataset.
			We copied the results of UIF + Linear and SenSR from \cite{sensei}.}
		\label{table4}
		\footnotesize
		\centering
		\begin{center}
			\begin{tabular}{l|cccc}
				\toprule
				\textit{Adult} & Acc &  BA & S-Con. & GR-Con. \\
				\midrule
				\midrule
				UIF + Hinge & 85.3 & 76.8 & .936 & .967  \\
				UIF + SLIDE & 85.1 & 76.6 & \underline{.970} & \textbf{.985}  \\
				UIF + HySLIDE & 85.2 & 76.6 &  \textbf{.976} & {.981} \\
				\midrule
				UIF + Linear (SenSeI) & - & 76.8 &  {.945} & .963  \\
				% SenSR + Explore & - & 79.4 & \underline{.966} & \textbf{.987} & \underline{.065} & .044 & \underline{.084} & .059 \\
				SenSR  & 78.7 & 78.9 & .934 & \underline{.984}  \\
				% Baseline & 81.3 & 82.9 & .848 & .865 & .179 & .089 & .216 & .105 \\
				% Project & 81.3 & 82.7 & .868 & 1.00 & .145 & .064 & .192 & .086 \\
				% Adv. debiasing & 81.2 & 81.5 & .807 & .841 & .082 & .070 & .110 & .078 \\
				% CoCL & & 79.0 & - & - & .163 & .080 & .201 & .109 \\
				\bottomrule
			\end{tabular}
		\end{center}
	\end{table}

	\section{Conclusion and discussion}
	\label{sec7}
	
	The main message of this paper is to show that using 
	%the hinge function as a surrogate function in the fairness constraint would yield suboptimal results.
	the hinge-surrogate constraint does not provide an optimal prediction model for a given fairness constraint.
	Thus, we need to be careful to choose a surrogate fairness constraint, and the choice should depend on the context of a given problem. For example, if the problem is fair classification, one can choose the proposed SLIDE-surrogate fairness constraint, which is theoretically and numerically reasonable.
	
	By closely investigating the gradient of the SLIDE-surrogate fairness constraint, we find an interesting new group fairness constraint so called the \textit{DI-boundary} defined as $\phi^{\textup{DI-bound}}(f)=|\mathbf{P}\{0\le f(\bX)\le \tau|Z=0\}-\mathbf{P}\{ 0\le f(\bX) \le \tau|Z=1\}|.$
	The DI-boundary requires fairness only for individuals whose scores are around the decision boundary. That is, when the fairness of $f$ is assessed, individuals who have very large scores (i.e., super-performers if a larger value of $f$ means a higher ability) are excluded from the analysis. If we replace the indicator function in $\phi^{\textup{DI-bound}}(f)$ by the hinge function, the gradient of the hinge-surrogate
	DI-boundary is equal proportional to the gradient of the SLIDE-surrogate DI provided that $0 \le f(\bX) \le \tau$ is replaced by $0 \le f^{\textup{curr}}(\bX) \le \tau,$ where $f^{\textup{curr}}$ is the current solution. Most existing fairness constraints require a prediction model fair for all individuals.
	It would be useful to think about the concept of partial fairness, where $f$ is fair only for a specific subset of the population.
	
	While we have focused on learning fair prediction models, 
	evaluation of the fairness of a given prediction model has also been received much attention
	(\cite{https://doi.org/10.48550/arxiv.2103.16714,10.1145/3442188.3445884}).
	Specially designed surrogate fairness constraints for the purpose of evaluation of the fairness would be necessary, which we leave as future work.
	
	%\section*{Acknowledgements}
	%
	%This work was supported by Institute for Information \& communications Technology Planning \& Evaluation(IITP) grant funded by the Korea government(MSIT) (No. 2019-0-01396, Development of framework for analyzing, detecting, mitigating of bias in AI model and training data).

	%%%%%%%%%%%%%%%%%%%%%%%%%%%%%%%%%%%%%%%%%%%%%%%%%%%%%%%%%%%%%%

	% Main text
	% \section{}\label{}
	
	% Use the style of numbering in square brackets.
	% If nothing is used, default style will be taken.
	%\begin{enumerate}[a)]
	%\item 
	%\item 
	%\item 
	%\end{enumerate}  
	
	%\begin{itemize}
	%\item 
	%\item 
	%\item 
	%\end{itemize}  
	
	%\begin{description}
	%\item[]
	%\item[] 
	%\item[] 
	%\end{description}  
	
	% Figure
	%\begin{figure}[<options>]
	%	\centering
	%		\includegraphics[<options>]{}
	%	  \caption{}\label{fig1}
	%\end{figure}

	% \begin{table}[<options>]
	% \begin{tabular*}{\tblwidth}{@{}LL@{}}
	% \toprule
	% \midrule
	%  & \\
	%  & \\
	%  & \\
	%  & \\
	% \bottomrule
	% \end{tabular*}
	% \end{table}
	
	% Uncomment and use as the case may be
	%\begin{theorem} 
	%\end{theorem}
	
	% Uncomment and use as the case may be
	%\begin{lemma} 
	%\end{lemma}
	
%	\clearpage
	
	\begin{center}
		{\Large \textbf{Appendix}}
	\end{center}
	
	\appendix
	
	In this Appendix, we present (i) the proofs of Theorems \ref{thm:fair_conv_di}, \ref{thm:fair_conv_uif} and \ref{thm:risk_conv_di} in Sections \ref{sec:A}, \ref{sec:B} and \ref{sec:C}, (ii) details of experiments in Section \ref{sec:D} and (iii) results of additional experiments in Section \ref{sec:E}.
	
	\section{Notations and definitions}\label{sec:A}
	
	Denote the compact domain of inputs as $\mathcal{X}\subset \mathbb{R}^d.$ 
	For a given function $f,$ we let $|| \cdot ||_{p}$ be the $l_p$ norm defined  as $||f||_{p} := \left( \int_{\mathcal{X}} | f(\mathbf{x}) |^{p} \,d\mu(\mathbf{x}) \right)^{1/p}$
	where $\mu$ is the Lebesgue measure on $\mathbb{R}^d.$ 
	In addition, we let $||f||_{\infty} := \sup_{\mathbf{x} \in \mathcal{X}} |f(\mathbf{x})|. $ 	
	
	For a given $\alpha>0,$ $[\alpha]$ is the largest integer less than or equal to $\alpha$ and
	$\lceil \alpha \rceil$ is the smallest integer greater or equal to $\alpha.$
	For given $\bfs = \left[ s_{1}, \cdots, s_{d} \right]^{\top} \in \mathbb{N}_0^d,$ where $\mathbb{N}_{0}$ is the set of nonnegative integers,
	we define the derivative of $f$ of order $\mathbf{s}$ as
	$$
	\partial^{ \mathbf{s} } f = \frac{  \partial^{|\mathbf{s}|} f }{ \partial x_{1}^{s_{1}} \cdots \partial x_{d}^{s_{d}} },
	$$
	where $| \mathbf{s} | = s_{1} + \cdots s_{d}.$
	Further, we let
	$$
	\left[ f \right]_{r, \mathcal{X}} = \sup_{\mathbf{x}, \mathbf{x}' \in \mathcal{X}, \mathbf{x} \neq \mathbf{x}'} \frac{| f(\mathbf{x}) - f(\mathbf{x}') |}{| \mathbf{x} - \mathbf{x}' |^{r}}
	$$
	for $r \in (0, 1].$
	
	\begin{definition}[Smooth functions]
		For $m \in \mathbb{N},$ we denote $ C^{m}(\mathcal{X}) $ as the space of $m$-times differentiable functions on $\mathcal{X}$ as whose partial derivatives of order $\mathbf{s}$ with $ | \mathbf{s} | \le m$ are continuous. That is,
		$$ C^{m}(\mathcal{X}) = \{  f : \mathcal{X} \rightarrow \mathbb{R}, $$
		$$ \partial^{\mathbf{s}} f \textup{ are continuous for } \forall \mathbf{s} \textup{ such that } | \mathbf{s} |  \le m \}. $$
	\end{definition}
	
	\begin{definition}[H\"{o}lder space]\label{holder}
		H\"{o}lder space with smoothness $\zeta > 0$ is a function space defined as
		$$
		\mathcal{H}^{\zeta}(\mathcal{X}) := \{ f \in C^{\left[ \zeta \right]}(\mathcal{X}) : ||f||_{\mathcal{H}^{\zeta} (\mathcal{X}) } < \infty \}
		$$
		where 
		$$ || f ||_{ \mathcal{H}^{\zeta} (\mathcal{X}) } = \max_{| \mathbf{m} | \le [ \zeta ] } || \partial^{\mathbf{m}} f ||_{\infty, \mathcal{X}} + \max_{| \mathbf{m} | = [\zeta] } [ \partial^{\mathbf{m}} f ]_{\zeta_{0}, \mathcal{X}}. $$
	\end{definition}
	
	\begin{definition}[The $\epsilon$-covering number]
		The $\epsilon$-covering number of a given class of functions $\mathcal{F}$ is the cardinality of the minimal $\epsilon$-covering set of $\mathcal{F}$ with respect to the $L_{p}$ norm, which is defined as:
		$$ N(\epsilon, \mathcal{F}, || \cdot ||_{p}) := $$
		$$ \inf \{ n \in \mathbb{N} : \exists f_{1}, \cdots, f_{n} \textup{ s.t. } \mathcal{F} \subset \bigcup_{i=1}^{n} B_{p}(f_{i}, \epsilon) \}, $$
		where $B_{p}(f_{i}, \epsilon) := \{f \in \mathcal{F} : || f - f_{i} ||_{p} \le \epsilon \}.$
	\end{definition}

	\begin{definition}[Metric entropy]
		The ($\epsilon$-) metric entropy of $\mathcal{F}$ (w.r.t. $L_{p}$ norm) is a logarithm of the $\epsilon$-covering number of $\mathcal{F}$ (w.r.t. $L_{p}$ norm), i.e., 
		$$ H(\epsilon, \mathcal{F}, || \cdot ||_{p}) := \log{( N(\epsilon, \mathcal{F}, || \cdot ||_{p}) )}.$$ 
	\end{definition}
	
	\begin{definition}[Rademacher complexity]
		Let $\sigma$ be a random variable having -1 or 1 with probability $1/2$ each. For independent realizations
		$\sigma_1,\ldots,\sigma_n$ of $\sigma,$ we define the empirical Rademacher complexity of a function class $\mathcal{F}$ as 
		$$ \hat{\mathcal{R}}(\mathcal{F}) := \frac{1}{n} \mathbf{E}_{\sigma} \left( \sup_{f \in \mathcal{F}} \sum_{i=1}^{n} \sigma_{i} f(\mathbf{X}_{i}) \right). $$
		The Rademacher complexity is the expectation of the empirical one with respect to $\mathbf{X}.$ That is, the (population) Rademacher complexity of $\mathcal{F}$ is
		$$ \mathcal{R}(\mathcal{F}) := \mathbf{E}_{\mathbf{X}} \left( \hat{\mathcal{R}} (\mathcal{F}) \right). $$
	\end{definition}

	\section{Technical lemmas}\label{sec:B}
	
	This section introduces technical lemmas used to prove the Theorems \ref{thm:fair_conv_di}, \ref{thm:fair_conv_uif} and \ref{thm:risk_conv_di}. 
	Particularly, Lemma \ref{lem:nn_rad_bound} provides a tight upper bound of $\hat\cR(\nu_{\tau}(\cF))$ when $\cF$ is class of deep
	neural networks, which plays a key role in deriving the convergence rates.
	In addition, even though we do not derive the convergence rates, we demonstrate how to
	obtain a tight upper bound of $\hat\cR(\nu_{\tau}(\cF))$ when $\cF$ is a class of linear functions in Example \ref{eg1}.
	For technical simplicity, we assume that the distribution of $\bX$ has a density $p(\bx)$ such that
	$0< \inf_{\bx\in \cX} p(\bx) \le \sup_{\bx\in \cX} p(\bx)<\infty.$

	\begin{lemma}
		\label{lem:uif_lip}
		Let $\eta_f(\bx) := D(f(\bx),f(\bx'))-\gamma$ with a Lipschitz (on bounded domain) metric $D(\cdot, \cdot)$ for a given $\gamma > 0.$ Then, there exists $c>0$ such that
		$$\|\eta_{f_1}-\eta_{f_2}\|_\infty \le c \|f_1-f_2\|_\infty$$
		for any two functions $f_1$ and $f_2.$
	\end{lemma}
	
	\begin{lemma}[Theorem 26.5 of \cite{shalev2014understanding}]
		\label{lem:rademacher}
		Let $\cH$ be a set of real-valued functions such that $\| l \circ h \|_\infty\le H$ for any $h\in\cH$ for some $H > 0.$ Then,
		\begin{equation*}
			L_P(h)\le L_n(h)+2\hat\cR_n( l \circ \cH)+4H\sqrt{\frac{2\log(4/\delta)}{n}}
		\end{equation*}
		for any $h\in\cH$, with probability at least $1-\delta > 0.$
	\end{lemma}

	\begin{lemma}[Dudley's Theorem (Theorem 1.19 of \cite{wolf2018mathematical})]
		\label{lem:dudley}
		Let $\cH$ be a set of real-valued functions such that $\|h\|_\infty\le H$ for any $h\in\cH$ for some $H>0$.  Then,
		\begin{equation}
			\hat\cR(\cH)\le \inf_{ \alpha \in [0, \frac{H}{2} ) }\left( 4\alpha + \frac{12}{\sqrt{n}}\int_{\alpha}^{H}\sqrt{\log \cN(\epsilon, \cH,  \|\cdot\|_{n,2})} d\epsilon\right)
		\end{equation}
		where $\|\cdot\|_{n,2}$ denotes the empirical $L_2$ norm defined by $\|h\|_{n,2}:=\sqrt{n^{-1}\sum_{i=1}^nh(\bX_i)^2}$ for $h\in\cH$.
	\end{lemma}
	
	\begin{lemma}[Lemma 3 of \cite{suzuki2018adaptivity}]
		\label{lem:covering-l0}
		Let $\cF$ be a set of  deep neural networks with the ReLU activation function, $L$ many layers, $N$ many nodes at each layer, $S$ many nonzero weights and biases that are bounded by $B$. Then for any $\epsilon>0$, 
		\begin{equation}
			\log \cN(\epsilon, \cF, \|\cdot\|_{\infty})
			\le 2S(L+1)\log \left(\frac{ (L+1)(N+1) B}{\epsilon}\right).
		\end{equation}
	\end{lemma}

	\begin{lemma}
		\label{lem:nn_rad_bound}
		Let $g:\mathbb{R}\to\mathbb{R}$ be a Lipschitz function in a sense that $|g(z_1)-g(z_2)|\le C'|z_1-z_2|$ for any $z_1,z_2\in\mathbb{R}$ for some constant $C'>0$. Let $\cF$ be a set of  deep neural networks with the ReLU activation function, $L$ many layers, $N$ many nodes at each layer, $S$ many nonzero weights and biases that are bounded by $B$. Let $g(\cF):=\{g(f):f\in\cF\}$ and $F'=\sup_{f\in\cF}\|g(f)\|_\infty$. Then
		\begin{equation*}
			\hat\cR(g(\cF))\le \frac{4}{n}+ \frac{12 F'}{\sqrt{n}}\sqrt{S(L+1)\log (C'n (L+1)(N+1) B)}.
		\end{equation*}
	\end{lemma}
	
	The next two propositions compute $M_{n, f}$ for UIF and DI.
	
	\begin{proposition}[Upper bound of $M_{f}$ for UIF]\label{prop1}
		Let $\cF$ be the class of DNN models considered in Theorem \ref{thm:risk_conv_di}.
		Suppose $\tau_n=\mathcal{O}(b_n),$ where $b_n$ is the sequence defined in Theorem \ref{thm:risk_conv_di}.
		Then we have
		$$ M_{f}  \le \frac{1}{\tau_{n}} | \phi_{n, -\tau_{n}}^{\textup{slide}}(f; \gamma, \epsilon) - \phi_{n, \tau_{n}}^{\textup{slide}}(f; \gamma, \epsilon) | + \mathcal{O}(1)$$
		for all $f \in \mathcal{F}.$
	\end{proposition}
	
	\begin{proposition}[Upper bound of $M_{f}$ for DI]\label{prop2}
		Let $\cF$ be the class of DNN models considered in Theorem \ref{thm:risk_conv_di}.
		Suppose $\tau_n=\mathcal{O}(b_n),$ where $b_n$ is the sequence defined in Theorem \ref{thm:risk_conv_di}.
		Then we have for all $f \in \mathcal{F},$
		\begin{equation*}
			\begin{split}
				M_{f} & \le \frac{1}{\tau_{n}} \sum_{z = 0, 1} \left| \frac{1}{n_{z}} \sum_{z_{i} = z} \left( \nu_{-\tau_{n}} (f(\mathbf{x}_{i})) - \nu_{\tau_{n}} (f(\mathbf{x}_{i})) \right) \right| \\
				& + \frac{1}{\tau_{n}} | \phi_{n, \tau_{n}}^{\textup{slide}}(f) - \phi_{n}(f) | \\
				& + \mathcal{O}(1).
			\end{split}
		\end{equation*}
		% 	$$ M_{f}  \le \frac{1}{\tau_{n}} \left( \sum_{z = 0, 1} \left| \frac{1}{n_{z}} \sum_{z_{i} = z} \left( \nu_{-\tau_{n}} (f(\mathbf{x}_{i})) - \nu_{\tau_{n}}^{\textup{slide}} (f(\mathbf{x}_{i})) \right) \right| \\
		% 	& + | \phi_{n, \tau_{n}}^{\textup{slide}}(f) - \phi_{n}(f) | \right) + \mathcal{O}(1)$$
		% 	for all $f \in \mathcal{F}.$
	\end{proposition}
	
	%	\begin{lemma}
	%		Consider any $\tau_{n} >0.$
	%		\begin{enumerate}
	%			\item (DI) Let $M_{f} := \sup_{r \in [0, \tau_{n}], z \in \{ 0, 1 \}} g_{f|z} (r) $
	%			where $g_{f|z}$ is the density of $f(\mathbf{X})$ conditional on $Z = z.$
	%			Then, for any $f \in \mathcal{F},$
	%			$$ | \phi_{\tau_{n}} (f) - \phi(f) | \le C M_{f} \tau_{n}$$ for some $C> 0.$
	%			
	%			\item (UIF) Let $D_{f} := \sup_{r \in [0, \tau_{n}]} g_{D} (r) $
	%			where $g_{D}$ is the density of $\sup_{\mathbf{v}: d( \mathbf{X}, \mathbf{v} ) \le \epsilon} D(f(\mathbf{X}), f(\mathbf{v})).$
	%			Then, for any $f \in \mathcal{F},$
	%			$$ | \phi_{\tau_{n}} (f; \gamma, \epsilon) - \phi(f ; \gamma, \epsilon) | \le C D_{f} \tau_{n}$$ for some $C> 0.$
	%		\end{enumerate}
	%	\end{lemma}
	
	Obtaining a tight upper bound of the Rademacher complexity plays a key role to derive a fast convergence rate.
	We need an upper bound of $\hat{\cR}(\nu_\tau(\cF))$ while that of $\hat{\cR}(\cF)$ is well known for many classes of $\cF.$
	Since $\nu_\tau$ is $1/\tau$-Lipschitz, we may use the inequality $\hat{\cR}(\nu_\tau(\cF))\le \hat{\cR}(\cF)/\tau$
	to derive an upper bound.  However, such a naive approach would not yield a good upper bound in particular when
	$\tau$ converges to 0 as $n$ increases. For DNNs, we derive an upper bound in Lemma \ref{lem:nn_rad_bound}
	by directly calculating the corresponding metric entropy. In the following example, we illustrate how to derive a good upper bound of the Rademacher complexity when $\cF$ is a class of linear functions. In particular, we derive an upper bound of the $L_2$-metric entropy
	of $\nu_\tau(\cF),$ with which we could obtain an upper bound of the Rademacher complexity by use of Lemma \ref{lem:dudley}.

	\begin{Example}[Linear model $\mathcal{F}^{\textup{linear}}$]\label{eg1}
		Let $\cF^{\textup{linear}}=\{\langle\bw,\cdot\rangle:\bw\in\mbR^d,\|\bw\|_2\le B\}.$
		Since $\nu_\tau$ is $1/\tau$-Lipschitz, we have
		$ \cN(\epsilon, \nu_\tau(\cF^{\textup{linear}} ),\|\cdot\|_2) \le \cN(\epsilon \tau, \cF^{\textup{linear}} ,\|\cdot\|_\infty).$
		Finally, we can obtain the bound of the entropy as 
		$ \log \mathcal{N}(\epsilon \tau_{n}, \mathcal{F}^{\textup{linear}},  || \cdot ||_{\infty}) \le d \log \left( \frac{C}{\epsilon \tau_n} \right) $
		for some $C > 0$ and $\tau = \tau_{n}$ by Lemma 2.5 of \cite{Geer2000EmpiricalPI}.
		Hence, the Rademacher complexity of $\nu_{\tau_{n}}(\mathcal{F}^{\textup{linear}})$ is bounded by (asymptotically) 
		$\log {(1 / \tau_{n})}$ but not $1 / \tau_{n}.$
	\end{Example}
	
	% C
	\section{Proofs}\label{sec:C}
	
	This section presents the proofs of Lemmas \ref{lem:uif_lip} and \ref{lem:nn_rad_bound}, Theorems \ref{thm:fair_conv_di}, \ref{thm:fair_conv_uif} and \ref{thm:risk_conv_di}, and Propositions \ref{prop1} and \ref{prop2}.
	
	\paragraph{Proof of Lemma \ref{lem:uif_lip}}
	
	\begin{proof}
		
		By the Lipschitz property of $D,$ there exists a constant $C > 0$ such that $D(z, z') \le C |z - z'|$ for all $z, z' \in \{ x \in \mathbb{R}: |x| \le K \},$ which is a bounded domain, for a given $K > 0.$
		
		Denote $$ \mathbf{x}_{f_{1}}' := \argmax_{\mathbf{x}' : d(\mathbf{x}, \mathbf{x}') \le \epsilon} D(f_{1}(\mathbf{x}), f_{1}(\mathbf{x}')) $$ and $$ \mathbf{x}_{f_{2}}' := \argmax_{\mathbf{x}' : d(\mathbf{x}, \mathbf{x}') \le \epsilon} D(f_{2}(\mathbf{x}), f_{2}(\mathbf{x}')). $$
		Then, we consider any metric as the similarity measure $D(\cdot, *)$ between prediction values, we can write
		$ | \eta_{f_{1}}(\mathbf{x}) - \eta_{f_{2}}(\mathbf{x}) | = | D ( f_{1}(\mathbf{x}), f_{1}(\mathbf{x}_{f_{1}}') ) - D ( f_{2}(\mathbf{x}), f_{2}(\mathbf{x}_{f_{2}}') ) |. $
		
		(Case 1) $D ( f_{1}(\mathbf{x}) , f_{1}(\mathbf{x}_{f_{1}}') ) \ge D ( f_{2}(\mathbf{x}) , f_{2}(\mathbf{x}_{f_{2}}') ):$ 
		Then,
		\begin{equation*}
			\begin{split}
				\left| \eta_{f_{1}}(\mathbf{x}) - \eta_{f_{2}}(\mathbf{x}) \right| & = D (  f_{1}(\mathbf{x}) , f_{1}(\mathbf{x}_{f_{1}}') ) - D ( f_{2}(\mathbf{x}) , f_{2}(\mathbf{x}_{f_{2}}') )
				\\
				& \le D (  f_{1}(\mathbf{x}) , f_{1}(\mathbf{x}_{f_{1}}') ) - D ( f_{2}(\mathbf{x}) , f_{2}(\mathbf{x}_{f_{1}}') )
				\\
				& \le \big| D (  f_{1}(\mathbf{x}) , f_{1}(\mathbf{x}_{f_{1}}') ) - D ( f_{2}(\mathbf{x}) , f_{2}(\mathbf{x}_{f_{1}}') ) \big|
				\\
				& \le D ( f_{1}(\mathbf{x}) , f_{2}(\mathbf{x}) ) + D ( f_{1}(\mathbf{x}_{f_{1}}') , f_{2}(\mathbf{x}_{f_{1}}') ).
			\end{split}
		\end{equation*}
		The first inequality holds since 
		$$ D ( f_{2}(\mathbf{x}) , f_{2}(\mathbf{x}_{f_{2}}') ) \ge D (  f_{2}(\mathbf{x}) , f_{2}(\mathbf{x}')  ) $$
		for all $\mathbf{x}'.$
		
		(Case 2) $D ( f_{1}(\mathbf{x}) , f_{1}(\mathbf{x}_{f_{1}}') ) \le D ( f_{2}(\mathbf{x}) , f_{2}(\mathbf{x}_{f_{2}}') ):$
		We can derive $| \eta_{f_{1}}(\mathbf{x}) - \eta_{f_{2}}(\mathbf{x})  | \le D ( f_{1}(\mathbf{x})  , f_{2}(\mathbf{x}) ) + D ( f_{1}(\mathbf{x}_{f_{2}}') , f_{2}(\mathbf{x}_{f_{2}}') )$ similarly. 
		
		Since the two inequalities hold for any $\mathbf{x}$ and $D$ is Lipschitz, we can take supremum for each hand-side:
		\begin{equation*}
			\begin{split}
				|| \eta_{f_{1}} - \eta_{f_{2}} ||_{\infty} & = \sup_{\mathbf{x}} | \eta_{f_{1}}(\mathbf{x}) - \eta_{f_{2}}(\mathbf{x}) | \\
				& \le 3 \sup_{\mathbf{x}} D(f_{1}(\mathbf{x}) , f_{2}(\mathbf{x}) ) \\
				& \le 3 C \sup_{\mathbf{x}} | f_{1}(\mathbf{x}) - f_{2}(\mathbf{x}) | \\
				& = c || f_{1} - f_{2} ||_{\infty}
			\end{split}
		\end{equation*}
		where $c = 3C.$
		Here, we note that the first inequality holds by the inequality
		$ | \eta_{f_{1}}(\mathbf{x}) - \eta_{f_{2}}(\mathbf{x}) | \le D( f_{1}(\mathbf{x}), f_{2}(\mathbf{x}) ) + D ( f_{1}(\mathbf{x}_{f_{1}}') , f_{1}(\mathbf{x}_{f_{1}}') ) + D ( f_{2}(\mathbf{x}_{f_{2}}') , f_{2}(\mathbf{x}_{f_{2}}') ) $ using (Case 1) and (Case 2).
		It finally implies the desired result that
		$$
		|| \eta_{f_{1}} - \eta_{f_{2}}  ||_{\infty} \le c || f_{1} - f_{2} ||_{\infty}
		$$
		holds for some $c > 0$ that only depends on the Lipschitz constant $C.$
		
	\end{proof}

	\paragraph{Proof of Lemma \ref{lem:nn_rad_bound}}
	
	\begin{proof}
		By the Lipschitz property of $g$, we have 
		$\|g(f_1)-g(f_2)\|_{n,2}^2=\frac{1}{n}\sum_{i=1}^n|g(f_1(\bx_i))-g(f_2(\bx_i))|^2\le (C')^2\|f_1-f_2\|^2_\infty$
		and thus
		$\cN(\epsilon, g(\cF),  \|\cdot\|_{n,2}) \le \cN(\epsilon/C', \cF,  \|\cdot\|_{\infty}).$
		Then by Lemma \ref{lem:covering-l0}, the integral term in Lemma \ref{lem:dudley} is bounded as
		$$ \int_{\alpha}^{F'}\sqrt{\log \cN(u/C', \cF,  \|\cdot\|_{\infty})} d \epsilon $$
		$$	\le {F'}\sqrt{\log \cN(\alpha/C', \cF,  \|\cdot\|_{\infty})} $$
		$$ \le {F'}\sqrt{2S(L+1)\log \left(\frac{C'(L+1)(N+1) B}{\alpha}\right)}. $$
		Taking $\alpha=1/n$ completes the proof.
	\end{proof}
	
	%	\paragraph{Proof of Lemma 6}
	%	
	%	\begin{proof}
	%	Consider DI at first.
	%	Then, we can show easily by triangle inequality and integration formula as below.
	%	\begin{equation*}
	%		\begin{split}
	%			| \phi_{\tau_{n}}(f) - \phi(f) | & = | | \mathbf{E} ( \nu_{\tau_{n}} (f(\mathbf{X})) | Z = 0)  - \mathbf{E} ( \nu_{\tau_{n}} (f(\mathbf{X})) | Z = 1 ) | - | \mathbf{P}(f(\mathbf{X})> 0 | Z = 0) - \mathbf{P}(f(\mathbf{X})> 0 | Z = 1)  | | \\
	%			& \le | \mathbf{E} ( \nu_{\tau_{n}} (f(\mathbf{X})) | Z = 0) - \mathbf{P}(f(\mathbf{X})> 0 | Z = 0) | + | \mathbf{E} ( \nu_{\tau_{n}} (f(\mathbf{X})) | Z = 1) - \mathbf{P}(f(\mathbf{X})> 0 | Z = 1) | \\
	%			& = \mathbf{P}(0 < f(\mathbf{X}) < \tau_{n} | Z = 0) + \mathbf{P}(0 < f(\mathbf{X}) < \tau_{n} | Z = 1) \\
	%			& \le \int_{0}^{\tau_{n}} g_{f | Z = 0}(r) \,dr + \int_{0}^{\tau_{n}} g_{f | Z = 1}(r) \,dr \\
	%			& \le 2 \tau_{n} \sup_{r \in [0, \tau_{n}], z = 0, 1} g_{f | z} (r).
	%		\end{split}
	%	\end{equation*}
	%	Since we define $M_{f} = \sup_{r \in [0, \tau_{n}], z = 0, 1} g_{f | z} (r),$ it implies the desired result.
	%
	%	For UIF, we only modify as
	%	$| \phi_{\tau_{n}} (f; \gamma, \epsilon) - \phi(f; \gamma, \epsilon) | = \mathbf{P}( 0 < \sup_{\mathbf{v}: d(\mathbf{X}, \mathbf{v}) < \epsilon} D(f(\mathbf{X}), f(\mathbf{v})) < \tau_{n}). $
	%	Then, one can similarly show the same result for UIF using $D_{f}$ instead of $M_{f}.$
	%	\end{proof}

	\paragraph{Proof of Theorem \ref{thm:fair_conv_di}}
	
	\begin{proof}
		Let 
		$$ \phi_{\tau_n}^{\textup{slide}}(f)=|\mathbf{E}\{\nu_{\tau_n}(f(\bX))|Z=0\}-\mathbf{E}\{\nu_{\tau_n}(f(\bX))|Z=1\}|. $$
		Then using the inequality $|a|-|b|\le|a-b|$, 
		we have that
		\begin{equation*}
			\begin{split}
				& \left| \phi_{n,\tau_n}^{\textup{slide}}(\hat{f}_{n}) - \phi(\hat{f}_{n}) \right| - | \phi_{\tau_n}^{\textup{slide}}( \hat{f}_{n} ) - \phi( \hat{f}_{n} ) | \\
				& \le | \phi_{n,\tau_n}^{\textup{slide}}(\hat{f}_{n}) - \phi_{\tau_n}^{\textup{slide}}( \hat{f}_{n} ) | \\
				& = \bigg| \big| \frac{1}{n_0}\sum_{i:z_i=0}\nu_{\tau_n}( \hat{f}_{n} (\bx_i)) - \frac{1}{n_1}\sum_{i:z_i=1}\nu_{\tau_n}( \hat{f}_{n} (\bx_i)) \big| \\
				& - \big| \mathbf{E}\{\nu_{\tau_n}(\hat{f}_{n}(\bX))|Z=0\} - \mathbf{E}\{\nu_{\tau_n}(\hat{f}_{n}(\bX))|Z=1\} \big| \bigg| \\
				& \le \left| \frac{1}{n_0}\sum_{i:z_i=0}\nu_{\tau_n}( \hat{f}_{n} (\bx_i)) - \mathbf{E}\{\nu_{\tau_n}(\hat{f}_{n}(\bX))|Z=0\} \right| \\
				& + \left| \frac{1}{n_1}\sum_{i:z_i=1}\nu_{\tau_n}( \hat{f}_{n} (\bx_i)) - \mathbf{E}\{\nu_{\tau_n}(\hat{f}_{n}(\bX))|Z=1\} \right| \\
				& = \sum_{z\in\{0,1\}} \left|\frac{1}{n_z}\sum_{i:z_i=z}\nu_{\tau_n}( \hat{f}_{n} (\bx_i))-\mathbf{E}\{\nu_{\tau_n}( \hat{f}_{n} (\bX))|Z=z\}\right|.
			\end{split}
		\end{equation*}	
		%	so that we have 
		%	$$
		%	\left|\phi_{n,\tau_n}^{\textup{slide}}(\hat{f}_{n})-\phi(\hat{f}_{n})\right|
		%	$$
		%	$$
		%	\le  \sum_{z\in\{0,1\}} \left|\frac{1}{n_z}\sum_{i:z_i=z}\nu_{\tau_n}( \hat{f}_{n} (\bx_i))-\mathbf{E}\{\nu_{\tau_n}( \hat{f}_{n} (\bX))|Z=z\}\right|
		%	+ |\phi_{\tau_n}^{\textup{slide}}( \hat{f}_{n} )-\phi( \hat{f}_{n} )|.		
		%	$$
		Since the SLIDE function $\nu_{\tau_n}$ is  bounded in $[0,1]$,  by   Lemma \ref{lem:rademacher} the first term of the right-hand side of the preceding display is bounded by
		\begin{align*}
			&\sum_{z\in\{0,1\}}\left(\hat\cR_z(\nu_{\tau_n}(\cF))+4\sqrt{\frac{2\log(8n)}{n_z}}\right)
		\end{align*}
		with probability at least $1-1/n$. 
		On the other hand, the second term of the right-hand side is bounded by $M_{\hat{f}_n} \tau_n$ by the definition of $M_f.$
		By the assumptions $n_1/n_0\to s\in(0,\infty)$, we have 
		\begin{align*}
			\phi(\hat{f}_n)&\le \phi_{n,\tau_n}^{\textup{slide}}(\hat{f}_n) + \left|\phi_{n,\tau_n}^{\textup{slide}}(\hat{f}_{n})-\phi(\hat{f}_{n})\right|\\
			&\le \alpha+ \delta_n+ C \left( \sum_{z\in\{0,1\}}\hat\cR_z(\nu_{\tau_n}(\cF)) + \sqrt{\frac{\log n}{n}} \right) \\
			& + M_{\hat{f}_{n}} \tau_{n}
		\end{align*}
		for some constant $C>0$ with probability at least $1-1/n$, which completes the proof.
	\end{proof}

	\paragraph{Proof of Theorem \ref{thm:fair_conv_uif}}
	
	\begin{proof}
		Let $\phi_{\tau_n}^{\textup{slide}}(f;\gamma,\epsilon)=\E\{\nu_{\tau_n}\circ \eta_f\}$. Since the SLIDE function $\nu_{\tau_n}$ is bounded in $[0,1]$, Lemma \ref{lem:rademacher} implies that
		$$ \left|\phi_{n,\tau_n}^{\textup{slide}}(\hat{f}_{n}; \gamma, \epsilon)-\phi(\hat{f}_{n}; \gamma, \epsilon)\right|	$$
		$$ \le \left|\phi_{n,\tau_n}^{\textup{slide}}(\hat{f}_{n};\gamma,\epsilon)-\phi_{\tau_n}^{\textup{slide}}(\hat{f}_{n};\gamma,\epsilon)\right| $$
		$$ + \left|\phi_{\tau_n}^{\textup{slide}}(\hat{f}_{n};\gamma,\epsilon)-\phi(\hat{f}_{n};\gamma,\epsilon)\right|	$$
		$$ \le 2\hat\cR(\nu_{\tau_n}\circ \eta(\cF))+4\sqrt{\frac{2\log (4n)}{n}} + M_{\hat{f}_{n}} \tau_n $$
		with probability at least $1-1/n$, which completes the proof.
	\end{proof}

	\paragraph{Proof of Theorem \ref{thm:risk_conv_di}}
	\begin{proof}
		We prove Theorem \ref{thm:risk_conv_di} for DI and UIF separately. 
		
		\textbf{(Case 1: DI)}
		Since $f_\alpha^\star$ is a H\"{o}lder smooth function with smoothness $\zeta>0$, by Theorem 5 of \cite{schmidt2020nonparametric}, for any sufficiently large $n$, there is a neural network $f_n^*\in\cF$ such that
		\begin{equation*}
			\|f_n^*-f^\star_\alpha\|_\infty\le C_1' n^{-\frac{\zeta}{2\zeta+d}}=:b'_n
		\end{equation*}
		for some absolute constant $C_1'>0$, provided that $L_n=L_0\log n$, $N_n=N_0n^{\frac{d}{2\zeta+d}}$, $S_n= S_0n^{\frac{d}{2\zeta+d}}\log n$, $B_n=1$ and $F_n=F_0\ge \|f^\star_\alpha\|_\infty$, where $L_0$, $N_0$, $S_0$ and $F_0$ are constants depending only on $d$ and $\zeta$. From now on, we assume that the set $\cF$ include all deep neural networks of the architectures satisfying the above conditions. 
		By the Lipschitz property of $\phi$, we have
		\begin{equation*}
			\phi(f_n^*)\le \phi(f^\star_\alpha) + L\|f_n^*-f^\star_\alpha\|_\infty\le \alpha+Mb'_n
		\end{equation*}
		i.e., $f_n^*\in\cF_{\alpha+Mb'_n}$. Define the event
		\begin{align*}
			\cE_n(\xi):=\left\{\{(Y_i,\bX_i)\}_{i=1}^n:\sup_{f\in\cF}\left|\phi_{n,\tau_n}^{\textup{slide}}(f)-\phi(f)\right|\le \xi\right\}
		\end{align*}
		for $\xi>0$.
		Let $$ \phi_{\tau_n}^{\textup{slide}}(f)=|\mathbf{E}(\nu_{\tau_n}(f(\bX))|Z=0)-\mathbf{E}(\nu_{\tau_n}(f(\bX))|Z=1)|. $$
		Then using the inequality $|a|-|b|\le|a-b| $, we get 
		$$
		\left|\phi_{n,\tau_n}^{\textup{slide}}(\hat{f}_{n})-\phi(\hat{f}_{n})\right|
		$$
		$$	\le \sum_{z\in\{0,1\}} \left|\frac{1}{n_z}\sum_{i:z_i=z}\nu_{\tau_n}( \hat{f}_{n} (\bx_i))-\mathbf{E} (\nu_{\tau_n}( \hat{f}_{n} (\bX))|Z=z ) \right| $$
		$$ + |\phi_{\tau_n}^{\textup{slide}}( \hat{f}_{n} )-\phi( \hat{f}_{n} )|. $$
		Since the SLIDE function $\nu_{\tau_n}$ is $1/\tau_n$-Lipschitz and bounded in $[0,1]$,  by   Lemmas \ref{lem:rademacher} and \ref{lem:nn_rad_bound}, the first term of the right-hand side of the preceding display is bounded as
		\begin{align*}
			\sum_{z\in\{0,1\}} \frac{8}{n_z}
		\end{align*}
		\begin{align*}
			+ \sum_{z\in\{0,1\}} \frac{24}{\sqrt{n_z}}\sqrt{S_n(L_n+1)\log \frac{n(L_n+1)(N_n+1) B_n}{\tau_n}}
		\end{align*}
		\begin{align*}
			& + \sum_{z\in\{0,1\}} 4\sqrt{\frac{2\log(4/\delta)}{n_z}} \\
			& \le C_2'\sum_{z\in\{0,1\}}\left(\frac{1}{n_z}+\sqrt{ \frac{n^{\frac{p}{2\zeta+p}}(\log n)^3}{n_{z}} }+\sqrt{\frac{\log(1/\delta)}{n_z}}\right)
		\end{align*}
		for some constant $C_2'>0$ with probability at least $1-2\delta$. 
		On the other hand, the second term of the right-hand side is bounded by $M_{\hat{f}_n} \tau_n$ by the definition of $M_f.$	
		Hence, by the assumptions $n_1/n_0\to s\in(0,\infty)$, we have by taking $\delta=1/(8n)$, 
		$$ \left|\phi_{n,\tau_n}^{\textup{slide}}(\hat{f}_{n}) - \phi(\hat{f}_{n})\right| $$
		$$ \le  C_3'\left(n^{-\frac{\zeta}{2\zeta+p}}(\log n)^{3/2}\right)
		= C_3'b_n + M_{\hat{f}_{n}} \tau_{n} $$
		for some constant $C_3'>0$ with probability at least $1-1/(4n)$.    
		That is, $\mathbf{P} \{ \cE_n(C_3'b_n + M_{\hat{f}_{n}} \tau_{n})^c \} \le 1/(4n).$
		
		Now we show that the convergence rate of $\cE(\hat{f}_n,f_n^*)$ on $\cE_n(C_3'b_n+  M_{\hat{f}_{n}} \tau_{n})$. 
		Firstly we note that on $\cE(C_3'b_n+  M_{\hat{f}_{n}} \tau_{n}),$ $f_n^*\in\cF_{n,\alpha+Mb_n'+C_3'b_n+  M_{\hat{f}_{n}} \tau_{n}, \tau_{n}}^{\textup{slide}}.$
		
		Thus, since $\hat{f}_n$ is the ERM over $\cF_{n,\alpha+\delta_n}^{\textup{slide}}$, where $\delta_n>C_2b_n>Mb_n'+C_3'b_n+  M_{\hat{f}_{n}} \tau_{n}$ for sufficiently large $C_2>0$ by assumption and $\mathbf{P} \{ \cE_n(C_3'b_n+  M_{\hat{f}_{n}} \tau_{n}) \} \ge1-1/(4n)$, we have
		\begin{align*}
			\cE(\hat{f}_n,f_n^*) & \le L(\hat{f}_n)-L(f_n^*)-L_n(\hat{f}_n)+L_n(f_n^*) \\
			& \le 2\cR(g(\cF))+ 4(\log2 +2F_0)\sqrt{\frac{2\log (8n)}{n}}
		\end{align*}
		with probability at least $1-1/(2n)$, where we denote $L(f)=\mathbf{E} \{ l(Y,f(X)) \}$. Here, we let $g(f):=l(Y,f(\bX))-l(Y,f_n^*(\bX))$ and $g(\cF):=\{g(f):f\in\cF\}$. The second inequality follows from the Rademacher complexity bound in Lemma \ref{lem:rademacher} with the fact that $|g(f)|\le (\log2 +2F_0)$ for any $f\in \cF$. Moreover, since  $|g(f_1)-g(f_2)|\le |f_1(\bX)-f_2(\bX)|$ for any $(\bX, Y)$ due to the Lipschitz property of the logistic loss $l$, by Lemma \ref{lem:nn_rad_bound}, the preceding display is further bounded by
		\begin{align*}
			\frac{1}{\sqrt{n}}\sqrt{n^{\frac{d}{2\zeta+d}}(\log n)^3}
			=n^{-\frac{\zeta}{2\zeta+d}}(\log n)^{3/2}=b_n
		\end{align*}
		up to some multiplicative constant. Therefore,  there is a constant $C_4'>0$ such that
		\begin{align*}
			& \mathbf{P} \{ \cE(\hat{f}_n, f^\star_\alpha)>C_4'b_n+  M_{\hat{f}_{n}} \tau_{n} \} \\
			& \le  \mathbf{P} \{ \cE(\hat{f}_n, f_n^*)>(C_4'/2)b_n+  M_{\hat{f}_{n}} \tau_{n} \} \\
			& + \mathbf{P} \{ \cE(f_n^*,f^\star_\alpha)> (C_4'/2)b_n+  M_{\hat{f}_{n}} \tau_{n} \} \\
			& \le \mathbf{P} \{ \{\cE(\hat{f}_n, f_n^*)>(C_4'/2)b_n\}\cap \cE_n(C_3'b_n+  M_{\hat{f}_{n}} \tau_{n}) \} \\
			& + \mathbf{P} \{ \cE_n(C_3'b_n+  M_{\hat{f}_{n}} \tau_{n})^c \} \\
			& + \mathbf{P} \{ \|f_n^*-f^\star_\alpha\|_\infty> (C_4'/2)b_n+  M_{\hat{f}_{n}} \tau_{n} \} \\
			&\le 1-n^{-1},
		\end{align*}
		which completes the proof of the first assertion.
		
		The second assertion follows from the fact that on $\cE_n(C_3'b_n+  M_{\hat{f}_{n}} \tau_{n})$, 
		\begin{equation*}
			\begin{split}
				\phi(\hat{f}_n) & \le \phi_{n,\tau_n}^{\textup{slide}}(\hat{f}_n)+ C_3'b_n+  M_{\hat{f}_{n}} \tau_{n} \\
				& \le \alpha+\delta_n+C_3'b_n+  M_{\hat{f}_{n}} \tau_{n}.
			\end{split}
		\end{equation*}

		\textbf{(Case 2: UIF)}
		The proof is almost the same as that of (Case 1: DI). The only difference is that we need to derive bounds of the Rademacher complexity $\hat\cR(\nu_{\tau_n}\circ \eta(\cF))$ and the term 
		$$ \left|\phi_{n,\tau_n}^{\textup{slide}}(\hat{f}_{n}; \gamma, \epsilon)-\phi(\hat{f}_{n}; \gamma, \epsilon)\right|. $$
		
		By Lemma \ref{lem:uif_lip},
		\begin{align*}
			|\nu_{\tau_n}\circ\eta_{f_1}(\bx)-\nu_{\tau_n}\circ\eta_{f_2}(\bx)|
			\le \frac{C_1'}{\tau_n}\|f_1-f_2\|_\infty
		\end{align*}
		for any $f_1,f_2$ and any $\bx$ for some $C_1'>0.$ 
		Therefore, we have that
		\begin{equation*}
			\begin{split}
				\hat\cR(\nu_{\tau_n}\circ \eta(\cF)) 
				& \le \frac{4}{n} + \\
				& \frac{12}{\sqrt{n}}\sqrt{S_n(L_n+1)\log \frac{n(L_n+1)(N_n+1)B_n}{\tau_{n}}}
				\\
				& \le C_2' n^{-\frac{\zeta}{2\zeta+d}}(\log n)^{3/2}=C_2'b_n	        
			\end{split}
		\end{equation*}
		for some constant $C_2'>0$.
		In turn, similarly to the proof of (Case 1: DI), we have
		\begin{align*}
			\left|\phi_{n,\tau_n}^{\textup{slide}}(\hat{f}_{n}; \gamma, \epsilon)-\phi(\hat{f}_{n}; \gamma, \epsilon) \right|
			& \le  2\hat\cR(\nu_{\tau_n}\circ \eta(\cF)) \\
			& + 4\sqrt{\frac{2\log (4n)}{n}} \\
			& + M_{\hat{f}_{n}} \tau_n \\
			& \le  M_{\hat{f}_{n}} \tau_n + C_4'b_n 
		\end{align*}
		with probability  $1-1/(4n)$ for some constant $C_4'>0$. Finally, we can complete the proof similarly to the proof of (Case 1: DI).
	\end{proof}

	\paragraph{Proof of Proposition \ref{prop1}}
	
	\begin{proof}
		
		Note that $| \phi(f; \gamma, \epsilon) - \phi_{\tau_{n}}^{\textup{slide}}(f; \gamma, \epsilon) | \le | \phi_{\tau_{n}}^{\textup{slide}}(f; \gamma, \epsilon) - \phi_{-\tau_{n}}^{\textup{slide}}(f; \gamma, \epsilon) |.$
		By triangle inequality, we obtain $$| \phi_{\tau_{n}}^{\textup{slide}}(f; \gamma, \epsilon) - \phi_{-\tau_{n}}^{\textup{slide}}(f; \gamma, \epsilon) | $$
		$$ \le |\phi_{\tau_{n}}^{\textup{slide}}(f; \gamma, \epsilon) - \phi_{n, \tau_{n}}^{\textup{slide}}(f; \gamma, \epsilon) | $$
		$$ + | \phi_{-\tau_{n}}^{\textup{slide}}(f; \gamma, \epsilon) - \phi_{n, -\tau_{n}}^{\textup{slide}}(f; \gamma, \epsilon) | $$
		$$ + | \phi_{n, \tau_{n}}^{\textup{slide}}(f; \gamma, \epsilon) - \phi_{n, -\tau_{n}}^{\textup{slide}}(f; \gamma, \epsilon) |.$$
		The first and second terms can be bounded as below using Lemmas \ref{lem:rademacher} and \ref{lem:nn_rad_bound} using the same arguments used in the proof of Theorem \ref{thm:risk_conv_di}.
		That is, since the SLIDE functions $\nu_{\tau_n}$ and $\nu_{-\tau_{n}}$ are $1/\tau_n$-Lipschitz and bounded in $[0,1]$, 
		the first two terms are bounded by
		\begin{align*}
			& C' \left( \frac{1}{n} + \frac{1}{\sqrt{n}}\sqrt{S_n(L_n+1)\log \frac{n(L_n+1)(N_n+1)B_n}{\tau_{n}}} \right) \\
			& + 4\sqrt{\frac{2\log (1/(2\delta))}{n}} \\
			& \le C \left( \frac{1}{n}+\frac{1}{\sqrt{n}}\sqrt{n^{\frac{p}{2\zeta+p}}(\log n)^3}+\sqrt{\frac{\log(1/\delta)}{n}} \right)
		\end{align*}
		for some constant $C', C >0$ with probability at least $1-2\delta$.
		Now let $b_{n} = n^{-\frac{\zeta}{2\zeta + d} (\log n)^{3/2}}$ up to constant with $\delta = 1 / 8 n$ as is done in Theorem \ref{thm:risk_conv_di}. Then we obtain that $$| \phi_{\tau_{n}}^{\textup{slide}}(f; \gamma, \epsilon) - \phi_{n, \tau_{n}}^{\textup{slide}}(f; \gamma, \epsilon) | = \mathcal{O}(b_{n})$$ and $$| \phi_{-\tau_{n}}^{\textup{slide}}(f; \gamma, \epsilon) - \phi_{n, -\tau_{n}}^{\textup{slide}}(f; \gamma, \epsilon) | = \mathcal{O}(b_{n}).$$
		Thus we conclude 
		$$| \phi_{\tau_{n}}^{\textup{slide}}(f; \gamma, \epsilon) - \phi_{-\tau_{n}}^{\textup{slide}}(f; \gamma, \epsilon) | $$
		$$	\le | \phi_{n, \tau_{n}}^{\textup{slide}}(f; \gamma, \epsilon) - \phi_{n, -\tau_{n}}^{\textup{slide}}(f; \gamma, \epsilon) | +  \mathcal{O} (b_{n})$$ 
		and dividing by $\tau_{n}$ completes the proof.
	\end{proof}
	
	\paragraph{Proof of Proposition \ref{prop2}}
	
	\begin{proof}
		
		By triangle inequality, $ | \phi(f) - \phi_{\tau_{n}}^{\textup{slide}}(f) | \le | \phi_{\tau_{n}}^{\textup{slide}}(f) - \phi_{n, \tau_{n}}^{\textup{slide}}(f) | + | \phi(f) - \phi_{n}(f) | + | \phi_{n, \tau_{n}}^{\textup{slide}}(f) - \phi_{n}(f) |. $
		The first term is bounded as below using Lemmas \ref{lem:rademacher} and \ref{lem:nn_rad_bound} by the same arguments used in the proof of Theorem \ref{thm:risk_conv_di}.
		That is, since the SLIDE function $\nu_{\tau_{n}}$ are $1 / \tau_{n}$-Lipschitz and bounded in $[0, 1],$ it is bounded by
		\begin{align*}
			& \sum_{z\in\{0,1\}} \frac{8}{n_z} \\
			& + \sum_{z\in\{0,1\}} \frac{24}{\sqrt{n_z}}\sqrt{S_n(L_n+1)\log \frac{n(L_n+1)(N_n+1) B_n)}{\tau_n}} \\
			& + \sum_{z\in\{0,1\}} 4\sqrt{\frac{2\log(4/\delta)}{n_z}}
		\end{align*}
		$$ \le C \sum_{z\in\{0,1\}}\left(\frac{1}{n_z}+\frac{1}{\sqrt{n_z}}\sqrt{n^{\frac{p}{2\zeta+p}}(\log n)^3}+\sqrt{\frac{\log(1/\delta)}{n_z}}\right) $$
		for some constant $C >0$ with probability at least $1-2\delta$.
		Now let $b_{n} = n^{-\frac{\zeta}{2\zeta + d} (\log n)^{3/2}}$ up to constant with $\delta = 1 / 8 n$ as is done in Theorem \ref{thm:risk_conv_di}. Then we obtain that $| \phi_{\tau_{n}}^{\textup{slide}}(f) - \phi_{n, \tau_{n}}^{\textup{slide}}(f) | =	 \mathcal{O}(b_{n}).$
		On the other hand, using the inequality $ \big| |a-b| - |c-d| \big| \le | a - c | + | b - d |,$
		the second term is bounded as 
		\begin{equation*}
			\begin{split}
				| \phi(f) - \phi_{n}(f) |  & = \big| | E^{0} - E^{1} | - | E^{0}_{n} - E^{1}_{n}  | \big| \\
				& \le | E^{0} - E^{0}_{n} | + | E^{1} - E^{1}_{n} |
			\end{split}
		\end{equation*}
		where $E^{z} = \mathbf{E} \{ \mathrm{I}(f(\mathbf{X}) > 0) | Z = z \} $ and $E^{z}_{n} = \frac{1}{n_{z}} \sum_{z_{i} = z} \mathrm{I}(f(\mathbf{x}_{i}) > 0) $ for $z = 0, 1.$	
		Here, we note that 
		$ | E^{z} - E^{z}_{n} | = \max{( E^{z} - E^{z}_{n}, E^{z}_{n} - E^{z} )}.$
		Let $E^{z}_{\tau_{n}} = \mathbf{E} \{ \nu_{\tau_{n}}(f(\mathbf{X})) | Z = z \} $ and $E^{z}_{n, \tau_{n}} = \frac{1}{n_{z}} \sum_{z_{i} = z} \nu_{\tau_{n}}(f(\mathbf{x}_{i})).$
		Then,
		\begin{equation*}
			\begin{split}
				E^{z} - E^{z}_{n} & \le E^{z}_{-\tau_{n}} - E^{z}_{n, \tau_{n}} \le | E^{z}_{-\tau_{n}} - E^{z}_{n, -\tau_{n}} | + | E^{z}_{n, -\tau_{n}} - E^{z}_{n, \tau_{n}} | \\
				& \le C_{1} b_{n} + | E^{z}_{n, -\tau_{n}} - E^{z}_{n, \tau_{n}} |     
			\end{split}
		\end{equation*}
		and
		\begin{equation*}
			\begin{split}
				E^{z}_{n} - E^{z} & \le E^{z}_{n, -\tau_{n}} - E^{z}_{\tau_{n}} \le | E^{z}_{n, -\tau_{n}} - E^{z}_{n, \tau_{n}} | + | E^{z}_{n, \tau_{n}} - E^{z}_{\tau_{n}} | \\
				& \le | E^{z}_{n, -\tau_{n}} - E^{z}_{n, \tau_{n}} | + C_{2} b_{n}     
			\end{split}
		\end{equation*}
		for some constant $C_{1}, C_{2} > 0$ by Lemmas \ref{lem:rademacher} and \ref{lem:nn_rad_bound} as is done in Theorem \ref{thm:risk_conv_di}.
		Thus we have
		\begin{equation*}
			\begin{split}
				| \phi(f) - \phi_{\tau_{n}}^{\textup{slide}}(f) | & \le C b_{n} + \sum_{z = 0, 1} | E^{z}_{n, -\tau_{n}} - E^{z}_{n, \tau_{n}} | \\
				& + | \phi_{n, \tau_{n}}^{\textup{slide}}(f) - \phi_{n}(f) | \\
				& = C b_{n} + \\
				& \sum_{z = 0, 1} \left| \frac{1}{n_{z}} \sum_{z_{i} = z} \left( \nu_{-\tau_{n}} (f(\mathbf{x}_{i})) - \nu_{\tau_{n}} (f(\mathbf{x}_{i})) \right) \right| \\
				& + | \phi_{n, \tau_{n}}^{\textup{slide}}(f) - \phi_{n}(f) |
			\end{split}
		\end{equation*}
		for some $C > 0$.  Dividing the above inequality by $\tau_{n}$ completes the proof.
		
	\end{proof}
	
	% D
	\section{Implementation details}\label{sec:D}
	
	We use a DNN model with one hidden layer of size 100 with the softmax function at the output layer.
	For the optimizers, the Adam optimizer (\cite{adam}) is used with scheduling the learning rates reducing by half at every 500 epochs. For the datasets, we describe the sample sizes and feature dimensions in Table \ref{tabled1}. 
	In addition, we introduce the hyperparameters in Table \ref{tabled1} with ``Epochs'' (the total number of iterations), ``Optimizer'' (the gradient descent optimization algorithm) and ``lr'' (the initial learning rate). We fix $\gamma$ in UIF at 0.01 and choose $\tau \sim U(0.01, 0.2)$ along with random initials.

	\begin{table}[h]
		\footnotesize
		\centering
		\caption{Description of datasets and corresponding hyperparameters.}
		\begin{center}
			\begin{tabular}{l|ccc}
				\toprule
				Dataset & \textit{Adult} & \textit{Bank} & \textit{Law} \\
				\midrule
				\midrule
				$n_{train} / n_{test}$ & 36177 / 9044 & 32950 / 8238 & 21240 / 5310 \\
				$d$ & 41 & 47 & 11 \\
				Epochs & 2000 & 2000 & 2000 \\
				Optimizer & Adam & Adam & Adam \\
				lr & 0.5 & 0.05 & 0.005 \\
				$\lambda$ & (0.01, 10) & (0.01, 50) & (0.01, 100) \\
				$\tau$ & $\sim U$(0.01, 0.2) & $\sim U$(0.01, 0.2) & $\sim U$(0.01, 0.2) \\
				$\gamma$ (UIF) & 0.01 & 0.01 & 0.01 \\
				\bottomrule
			\end{tabular}
		\end{center}
		\label{tabled1}
	\end{table}

	\paragraph{Evaluation measures}
	The balanced accuracy of a trained prediction model $\hat{f}$ is $ \big( \sum_{i: y_{i} = -1} \mathrm{I}( \hat{f}(\mathbf{x}_{i}) = y_{i} )/n_{-1} + \sum_{i: y_{i} = 1} \mathrm{I}( \hat{f}(\mathbf{x}_{i}) = y_{i} )/n_1 \big) / 2. $ For the consistency (Con.), we compute the rate of predictions that do not change when only the sensitive variable is changed. That is, Con. is computed as $ \sum_{i=1}^{n} \mathrm{I}\{ \hat{f}(\mathbf{x}_{i, z_i=0}) = \hat{f}(\mathbf{x}_{i,z_i=1}) \} / n,$ 
	where $\mathbf{x}_{i,z_{i} = z}$ is an input vector which is the same as $\bx_i$ except $z_i=z.$ 
	Con. is initially considered by \cite{sensei, Yurochkin2020Training}. The S-Con. and GR-Con. in Table \ref{table3} of Section \ref{sec5} are the Con. values when $z$ is the ``spouse'' variable and the multiples of ``gender'' and ``race'' variables, respectively.

	\paragraph{Generating adversarial inputs for UIF}
	For UIF, in practice, we should compute an adversarial input $\mathbf{\bx}_{adv} := \argmax_{ \mathbf{x}' : d(\mathbf{x}, \mathbf{x}') \le \epsilon} D( f(\mathbf{x}), f(\mathbf{x}') ) $ of an arbitrary input $\mathbf{x}.$ We use $\mathbf{x}_{adv} := \mathbf{x} + \mathbf{r}_{adv},$ where $ \mathbf{r}_{adv}$ is an adversarial direction. It is approximated by the second order Taylor's polynomial with an approximated Hessian matrix as is proposed in VAT \cite{vat}. The source code that we use to generate $\mathbf{x}_{adv}$ is a modified version of the source code from \url{https://github.com/lyakaap/VAT-pytorch/blob/master/vat.py}, where we replace the Kullback-Leibler divergence in VAT by
	the similarity metric $D.$

	\section{Additional experiments}\label{sec:E}
	
	In this section, we present the results of additional experiments.
	
	\paragraph{An additional experiment similar to Figure \ref{fig2}}
	
\begin{figure}[h]
		\centering
		\includegraphics[scale = 0.155]{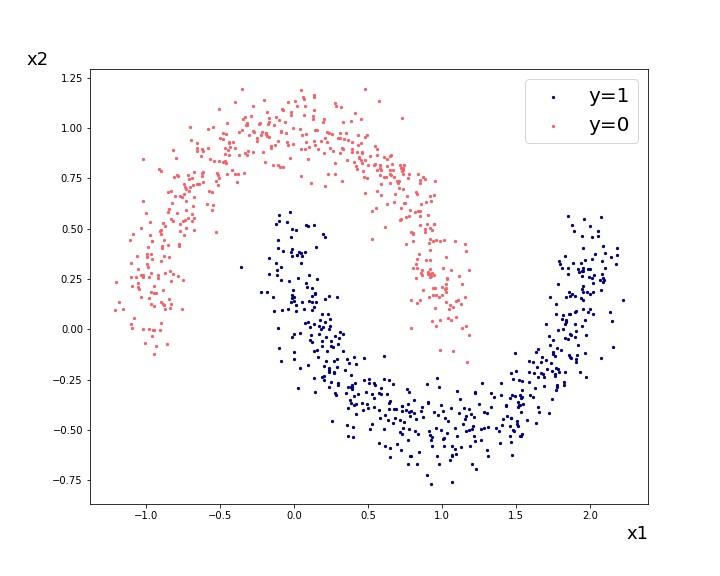}
		\includegraphics[scale = 0.235]{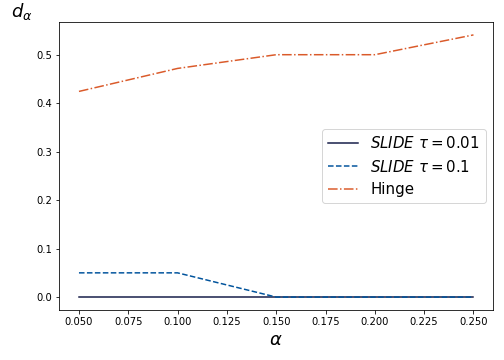}
		\caption{(Left) The two-moon dataset. (Right) Plot of $\alpha$ vs. $d_{\alpha}^{\textup{hinge}}$
			and $d_{\alpha,\tau}^{\textup{slide}}$ for $\tau\in \{0.01,0.1\}.$
		}
		\label{fig2-1}
	\end{figure}

	The 2-D dataset used in Figure \ref{fig2} is generated from two Gaussians: 
	$\mathbf{X} | Y = 0 \sim \mathcal{N}([0.5, 4.5]^{\top}, 2I_{2}), \mathbf{X} | Y = 1 \sim \mathcal{N}([2.0, 0.5]^{\top}, 2I_{2})$.
	We did a similar experiment with the two-moon dataset illustrated in the left panel of the Figure \ref{fig2-1}. 
	Even though the two used datasets in Figure \ref{fig2} and Figure \ref{fig2-1} are completely different, the behaviors of the Hausdorff distances are similar in the sense
	that the hinge-surrogate fairness constraint does not approximate the original fairness constraint well while the SLIDE works well.
	Note that we consider $\alpha$ less than 0.25 since the level of fairness for the optimal classifier is around 0.25 in this two-moon dataset.
	
	\paragraph{An example of inconsistency of fairness for the hinge-surrogate fairness constraint}
	
\begin{figure}[h]
		\centering
		\includegraphics[scale = 0.32]{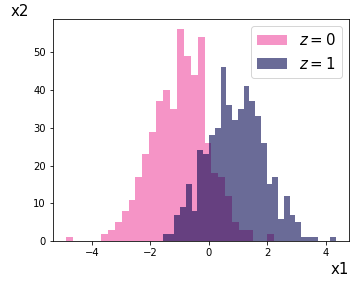}
		\includegraphics[scale = 0.325]{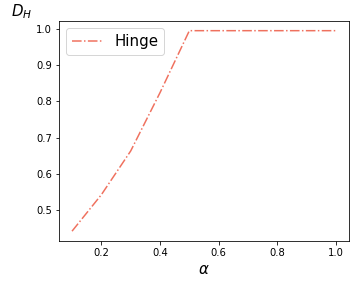}
		\caption{(Left) The generated dataset, $X|Z=z \sim \mathcal{N}(z,1).$ (Right) Plot of $\alpha$ vs. $d_{\alpha}^{\textup{hinge}}.$ 
		}
		\label{fige2}
	\end{figure}

	Let $X \in \mathbb{R}$ be a random variable having the distribution as $X | Z=z \sim \mathcal{N}(z,1)$ for $z\in \{-1,1\}.$
	For $\mathcal{F},$ we consider linear models as $\mathcal{F} = $ {$\beta_0 + \beta x: \beta_0\in [-1,1], \beta\in [-1,1]$}.
	Note that $\beta_0 + \beta X | Z=z \sim \mathcal{N}(\beta_0+\beta z,\beta^2).$ 
	Hence, the DI value for given $(\beta_0,\beta)$ is computed by
	$ \textup{DI}(\beta_0,\beta)= |\Phi(- \beta_0/\beta+1)-\Phi(-\beta_0/\beta-1) |,$
	where $\Phi(\cdot)$ is the cumulative distribution of $\mathcal{N}(0,1).$
	On the other hand, by formula of the mean of the truncated normal distribution, we have 
	$ 
	\mathbf{E} ( (\beta_0 + \beta \mathbf{X}+1)_+ | Z=z ) = \beta_0 + \beta z + 1 + \frac{ \phi(- (\beta_0 + \beta z + 1) / \beta ) }{1 - \Phi(-(\beta_0 + \beta z + 1) / \beta)} 
	$
	where $\phi(\cdot)$ is the probability density function of $\mathcal{N}(0, 1).$
	Thus we have
	$ \textup{DI}^{\textup{hinge}}(\beta_0,\beta)
	= \left| -2 \beta + \frac{ \phi(- (\beta_0 - \beta + 1) / \beta ) }{1 - \Phi(-(\beta_0 - \beta + 1) / \beta)} - \frac{ \phi(- (\beta_0 + \beta + 1) / \beta ) }{1 - \Phi(-(\beta_0 + \beta + 1) / \beta)} \right|.
	$
	The left panel of Figure \ref{fige2} draws the $d_{\alpha}^{\textup{hinge}}$
	from $\textup{DI}^{\textup{hinge}}(\beta_0,\beta)$ above, which clearly shows that the hinge-surrogate fairness constraint is not consistent in group fairness.
	
	\paragraph{About $M_{n, \hat{f}_{n}}$}
	
	Table \ref{table8} presents the values of $M_{n, \hat{f}_{n}} \tau_n /\phi( \hat{f}_{n})$ for the learned prediction models on the three datasets and two fairness criteria.
	Note that $M_{n, \hat{f}_{n}} \tau_n$ is not much larger  than 10\% of $\phi( \hat{f}_{n})$ for all the cases, which confirms the validity of the SLIDE-surrogate fairness constraint.
	
	\begin{table}[h]
		\footnotesize
		\centering
		\caption{Average (s.e.) values $M_{n, \hat{f}_{n}} \tau_{n} / \phi(\hat{f}_{n}) \times 100\%$ of the learned fair prediction models $\hat{f}_{n}.$ 
			We take the averages on random initial model parameters.}
		\begin{center}
			\begin{tabular}{l|c|c}
				\toprule
				Dataset & DI + SLIDE & UIF + SLIDE \\
				\midrule
				\midrule
				\textit{Adult} & 5.42 (4.73) & 10.65 (3.69) \\
				\textit{Bank} & 10.12 (4.90)& 2.67 (0.38) \\
				\textit{Law} & 8.17 (2.21) & 8.33 (0.82) \\
				\bottomrule
			\end{tabular}
		\end{center}
		\label{table8}
	\end{table}

	\paragraph{Simulation for Theorem \ref{thm:risk_conv_di}}
	
\begin{figure}[h]
		\centering
		\includegraphics[scale = 0.4]{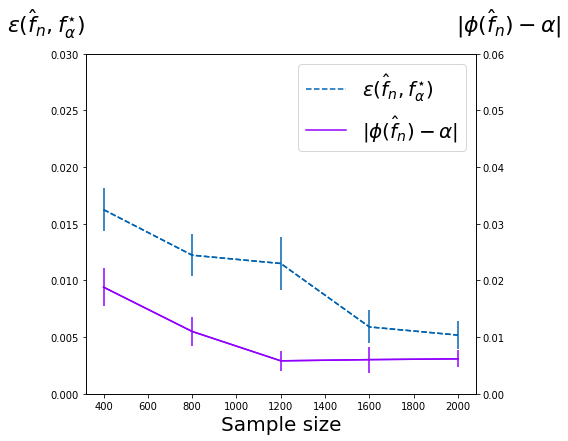}
		\caption{The $l$-excess risk $\mathcal{E}(\hat{f}_{n}, f_{\alpha}^{\star})$ and the fairness deviation $|\phi(\hat{f}_{n}) - \alpha|$
			for various sample sizes of the training data.}
		\label{fige4}
	\end{figure}

	We perform a numerical simulation supporting Theorem \ref{thm:risk_conv_di} in Section \ref{sec4}, which
	proves that the $l-$excess risk $\mathcal{E}(\hat{f}_{n}, f_{\alpha}^{\star})$ and
	the fairness risk $\phi(\hat{f}_{n}) - \alpha$  converge to 0 at certain rates as the sample size increases.
	For simulation data, consider the following distribution:
	$\mathbf{X} | S = 0, Y = 0 \sim \mathcal{N}(-1, 1.5),$
	$\mathbf{X} | S = 0, Y = 1 \sim \mathcal{N}(1.5, 0.5),$
	$\mathbf{X} | S = 1, Y = 0 \sim \mathcal{N}(-0.5, 1.0),$
	and
	$\mathbf{X} | S = 1, Y = 1 \sim \mathcal{N}(2.5, 1.5).$
	We choose the DI for the fairness criterion and the linear model for the classifier.
	% For the fairness criteria, we choose the DI, and for the prediction model, we consider the linear model.
	We set $\alpha = 0.2$ and  learn a fair prediction model $\hat{f}_{n}$ with the SLIDE-surrogate constraint (i.e., DI + SLIDE) with $\tau = 0.1.$ 
	Figure \ref{fige4} shows the excess risk $\mathcal{E}(\hat{f}_{n}, f_{\alpha}^{\star})$ and fairness deviation $|\phi(\hat{f}_{n}) - \alpha|$ for various sizes of training data. 
	It is obvious that the two quantities become smaller as the sample size increases, which clearly confirms the validity of the results of Theorem \ref{thm:risk_conv_di}.
	
	\begin{table}[h]
		\footnotesize
		\centering
		\caption{Comparison of UIF + $\{$Hinge, SLIDE, HySLIDE, SLIDE (CCCP)$\}$ on \textit{Adult}, \textit{Bank} and \textit{Law} test datasets.}
		\begin{center}
			\begin{tabular}{l|ccc}
				\toprule
				\textit{Adult} & Acc & BA & Con. (se) \\
				\midrule
				\midrule
				UIF + Hinge & 84.60 & 75.75 & .918 (.003) \\
				UIF + SLIDE & 84.36 & 75.73 & .920 (.005) \\
				UIF + HySLIDE & 84.51 & 75.69 & \textbf{.922} (.003) \\
				UIF + SLIDE (CCCP) & 84.52 & 75.70 & .921 (.004) \\
				\bottomrule
			\end{tabular}
			\begin{tabular}{l|ccc}
				\toprule
				\textit{Bank} & Acc & BA & Con. (se) \\
				\midrule
				\midrule
				UIF + Hinge & 90.29 & 63.82 & .985 (.006) \\
				UIF + SLIDE & 90.14 & 63.43 & \textbf{.990} (.005) \\
				UIF + HySLIDE & 90.15 & 63.03 & .984 (.004) \\
				UIF + SLIDE (CCCP) & 90.73 & 63.11 & .985 (.007) \\
				\bottomrule
			\end{tabular}
			\begin{tabular}{l|ccc}
				\toprule
				\textit{Law} & Acc & BA & Con. (se) \\
				\midrule
				\midrule
				UIF + Hinge & 83.75 & 62.67 & .985 (.006) \\
				UIF + SLIDE & 83.99 & 62.71 & .986 (.004) \\
				UIF + HySLIDE & 84.02 & 62.79 & .987 (.003) \\
				UIF + SLIDE (CCCP) & 84.11 & 62.66 & \textbf{.992} (.006) \\
				\bottomrule
			\end{tabular}
		\end{center}
		\label{tablee1}
	\end{table}

	\begin{table}[h]
		\footnotesize
		\centering
		\caption{Comparison of UIF + $\{$Hinge, SLIDE, HySLIDE, SLIDE (CCCP)$\}$ on \textit{Adult} test dataset. For UIF + Linear and SenSR, we copied the results from \cite{sensei}.}
		\begin{center}
			\begin{tabular}{l|cccc}
				\toprule
				\textit{Adult} & Acc &  BA & S-Con. & GR-Con.  \\
				\midrule
				\midrule
				UIF + Hinge & 85.3 & 76.8 & .936 & .967  \\
				UIF + SLIDE & 85.1 & 76.6 & {.970} & \textbf{.985}  \\
				UIF + HySLIDE & 85.2 & 76.6 & \textbf{.976} & {.981}  \\
				UIF + SLIDE (CCCP) & 85.2 & 76.8 & \underline{.974} & {.968}  \\
				\midrule
				UIF + Linear (SenSeI) & - & 76.8 & .945 & .963  \\
				SenSR & 78.7 & 78.9 & .934 & \underline{.984} \\
				\bottomrule 
			\end{tabular}
		\end{center}
		\label{tablee2}
	\end{table}
	
	\paragraph{Application of CCCP to UIF + SLIDE}
	
	We can apply the CCCP algorithm \cite{cccp} to find a local solution to the UIF + SLIDE constrained empirical risk minimization problem.
	The advantage of the CCCP algorithm is that it always converges to a local minimum.
	Note that the SLIDE function $\nu_\tau$ can be decomposed to the sum of the convex function
	$\nu_{\textup{conv}}(z)=(z)_+/\tau$ and the concave function $\nu_{\textup{conc}}=-(z-\tau)_+/\tau.$
	Note that we are to find a (local) minimum of $L_n(f)+\lambda \phi_{n,\tau}^{\textup{slide}}(f;\gamma,\epsilon).$
	Let $f^{\textup{curr}}$ be the current solution. Then, the CCCP algorithm updates $f$ by minimizing
	$$L_n(f)+\lambda \left\{ \phi_{n,\tau}^{\textup{slide, conv}}(f;\gamma,\epsilon) + f^\top 
	\nabla  \phi_{n,\tau}^{\textup{slide, conc}}(f^{\textup{curr}};\gamma,\epsilon)\right\} ,$$
	where
	$$\phi_{n,\tau}^{\textup{slide, conv}}(f;\gamma,\epsilon)=\frac{1}{n}\sum_{i=1}^n \left\{ D(f(\bx_i),f(\bv_i))-\gamma\right\}_+/\tau$$
	and 
	$$\phi_{n,\tau}^{\textup{slide, conc}}(f;\gamma,\epsilon)=- \frac{1}{n} \sum_{i=1}^n \left\{ D(f(\bx_i),f(\bv_i))-\gamma-\tau\right\}_+/\tau,$$
	and $\mathbf{v}_{i}$ is the adversarial input of $\mathbf{x}_{i},$ that is, 
	$$ \mathbf{v}_{i} = \argmax_{\mathbf{v} ; d(\mathbf{x}_{i}, \mathbf{v}) \le \epsilon} D(f( \mathbf{x}_{i} ), f( \mathbf{v} )). $$
	
	% \begin{table}[h]
	% 	\footnotesize
	% 	\centering
	% 	\caption{Comparison of UIF + $\{$Hinge, SLIDE, HySLIDE, SLIDE (CCCP)$\}$ on  \textit{Adult}, \textit{Bank} and \textit{Law} test datasets.}
	% 	\begin{center}
	% 		\begin{tabular}{c|c|cccc|}
	% 			\toprule
	% 			Dataset & UIF & + Hinge & + SLIDE & + HySLIDE & + SLIDE (CCCP) \\
	% 			\midrule
	% 			\multirow{3}{*}{\textit{Adult}} & Acc & 84.60 & 84.36 & 84.51 & 84.52 \\
	% 			& BA & 75.75 & 75.73 & 75.69 & 75.70 \\
	% 			& Con. (se) & .918 (.003) & .920 (.005) & \textbf{.922} (.003) & .921 (.004) \\
	% 			\midrule
	% 			\multirow{3}{*}{\textit{Bank}} & Acc & 90.29 & 90.14 & 90.15 & 90.73 \\
	% 			& BA & 63.82 & 63.43 & 63.03 & 63.11 \\
	% 			& Con. (se) & .985 (.006) & \textbf{.990} (.005) & .984 (.004) & .985 (.007) \\
	% 			\midrule
	% 			\multirow{3}{*}{\textit{Law}} & Acc & 83.75 & 83.99 & 84.02 & 84.11 \\
	% 			& BA & 62.67 & 62.71 & 62.79 & 62.66 \\
	% 			& Con. (se) & .985 (.006) & .986 (.004) & .987 (.003) & \textbf{.992} (.006) \\
	% 			\bottomrule
	% 		\end{tabular}
	% 	\end{center}
	% \end{table}

	Table \ref{tablee1} and Table \ref{tablee2} are the reproductions of Table \ref{table2} and Table \ref{table3} in the main paper with adding the results of the CCCP algorithm.
	The results suggest that the CCCP algorithm with the SLIDE is a promising alternative to the standard gradient descent algorithm.

	\printcredits
	
	\bibliographystyle{cas-model2-names}
	
	\bibliography{slide.bib}

\begin{thebibliography}{54}
\expandafter\ifx\csname natexlab\endcsname\relax\def\natexlab#1{#1}\fi
\providecommand{\url}[1]{\texttt{#1}}
\providecommand{\href}[2]{#2}
\providecommand{\path}[1]{#1}
\providecommand{\DOIprefix}{doi:}
\providecommand{\ArXivprefix}{arXiv:}
\providecommand{\URLprefix}{URL: }
\providecommand{\Pubmedprefix}{pmid:}
\providecommand{\doi}[1]{\href{http://dx.doi.org/#1}{\path{#1}}}
\providecommand{\Pubmed}[1]{\href{pmid:#1}{\path{#1}}}
\providecommand{\bibinfo}[2]{#2}
\ifx\xfnm\relax \def\xfnm[#1]{\unskip,\space#1}\fi
%Type = Inproceedings
\bibitem[{Agarwal et~al.(2018)Agarwal, Beygelzimer, Dudik, Langford and
  Wallach}]{pmlr-v80-agarwal18a}
\bibinfo{author}{Agarwal, A.}, \bibinfo{author}{Beygelzimer, A.},
  \bibinfo{author}{Dudik, M.}, \bibinfo{author}{Langford, J.},
  \bibinfo{author}{Wallach, H.}, \bibinfo{year}{2018}.
\newblock \bibinfo{title}{A reductions approach to fair classification}, in:
  \bibinfo{editor}{Dy, J.}, \bibinfo{editor}{Krause, A.} (Eds.),
  \bibinfo{booktitle}{Proceedings of the 35th International Conference on
  Machine Learning}, \bibinfo{publisher}{PMLR}. pp. \bibinfo{pages}{60--69}.
\newblock \URLprefix \url{https://proceedings.mlr.press/v80/agarwal18a.html}.
%Type = Article
\bibitem[{Angwin et~al.(2016)Angwin, Larson, Mattu and
  Kirchner}]{angwin2016machine}
\bibinfo{author}{Angwin, J.}, \bibinfo{author}{Larson, J.},
  \bibinfo{author}{Mattu, S.}, \bibinfo{author}{Kirchner, L.},
  \bibinfo{year}{2016}.
\newblock \bibinfo{title}{Machine bias}.
\newblock \bibinfo{journal}{ProPublica, May} \bibinfo{volume}{23},
  \bibinfo{pages}{2016}.
%Type = Inproceedings
\bibitem[{Awasthi et~al.(2021)Awasthi, Beutel, Kleindessner, Morgenstern and
  Wang}]{10.1145/3442188.3445884}
\bibinfo{author}{Awasthi, P.}, \bibinfo{author}{Beutel, A.},
  \bibinfo{author}{Kleindessner, M.}, \bibinfo{author}{Morgenstern, J.},
  \bibinfo{author}{Wang, X.}, \bibinfo{year}{2021}.
\newblock \bibinfo{title}{Evaluating fairness of machine learning models under
  uncertain and incomplete information}, in: \bibinfo{booktitle}{Proceedings of
  the 2021 ACM Conference on Fairness, Accountability, and Transparency},
  \bibinfo{publisher}{Association for Computing Machinery},
  \bibinfo{address}{New York, NY, USA}. p. \bibinfo{pages}{206–214}.
\newblock \URLprefix \url{https://doi.org/10.1145/3442188.3445884},
  \DOIprefix\doi{10.1145/3442188.3445884}.
%Type = Article
\bibitem[{Barocas and Selbst(2016)}]{diref}
\bibinfo{author}{Barocas, S.}, \bibinfo{author}{Selbst, A.D.},
  \bibinfo{year}{2016}.
\newblock \bibinfo{title}{Big data's disparate impact}.
\newblock \bibinfo{journal}{California Law Review} \bibinfo{volume}{104},
  \bibinfo{pages}{671--732}.
\newblock \URLprefix \url{http://www.jstor.org/stable/24758720}.
%Type = Article
\bibitem[{Bartlett et~al.(2006)Bartlett, Jordan and
  McAuliffe}]{bartlett2006convexity}
\bibinfo{author}{Bartlett, P.L.}, \bibinfo{author}{Jordan, M.I.},
  \bibinfo{author}{McAuliffe, J.D.}, \bibinfo{year}{2006}.
\newblock \bibinfo{title}{Convexity, classification, and risk bounds}.
\newblock \bibinfo{journal}{Journal of the American Statistical Association}
  \bibinfo{volume}{101}, \bibinfo{pages}{138--156}.
%Type = Article
\bibitem[{Blanchard et~al.(2008)Blanchard, Bousquet and
  Massart}]{blanchard2008statistical}
\bibinfo{author}{Blanchard, G.}, \bibinfo{author}{Bousquet, O.},
  \bibinfo{author}{Massart, P.}, \bibinfo{year}{2008}.
\newblock \bibinfo{title}{Statistical performance of support vector machines}.
\newblock \bibinfo{journal}{The Annals of Statistics} \bibinfo{volume}{36},
  \bibinfo{pages}{489--531}.
%Type = Inproceedings
\bibitem[{Calders et~al.(2009)Calders, Kamiran and
  Pechenizkiy}]{calders2009building}
\bibinfo{author}{Calders, T.}, \bibinfo{author}{Kamiran, F.},
  \bibinfo{author}{Pechenizkiy, M.}, \bibinfo{year}{2009}.
\newblock \bibinfo{title}{Building classifiers with independency constraints},
  in: \bibinfo{booktitle}{2009 IEEE International Conference on Data Mining
  Workshops}, \bibinfo{organization}{IEEE}. pp. \bibinfo{pages}{13--18}.
%Type = Inproceedings
\bibitem[{Celis et~al.(2019)Celis, Huang, Keswani and
  Vishnoi}]{celis2019classification}
\bibinfo{author}{Celis, L.E.}, \bibinfo{author}{Huang, L.},
  \bibinfo{author}{Keswani, V.}, \bibinfo{author}{Vishnoi, N.K.},
  \bibinfo{year}{2019}.
\newblock \bibinfo{title}{Classification with fairness constraints: A
  meta-algorithm with provable guarantees}, in: \bibinfo{booktitle}{Proceedings
  of the Conference on Fairness, Accountability, and Transparency}, pp.
  \bibinfo{pages}{319--328}.
%Type = Inproceedings
\bibitem[{Cho et~al.(2020)Cho, Suh and Hwang}]{cho2020fair}
\bibinfo{author}{Cho, J.}, \bibinfo{author}{Suh, C.}, \bibinfo{author}{Hwang,
  G.}, \bibinfo{year}{2020}.
\newblock \bibinfo{title}{A fair classifier using kernel density estimation},
  in: \bibinfo{booktitle}{34th Conference on Neural Information Processing
  Systems, NeurIPS 2020}, \bibinfo{organization}{Conference on Neural
  Information Processing Systems}.
%Type = Inproceedings
\bibitem[{Chuang and Mroueh(2021)}]{chuang2021fair}
\bibinfo{author}{Chuang, C.Y.}, \bibinfo{author}{Mroueh, Y.},
  \bibinfo{year}{2021}.
\newblock \bibinfo{title}{Fair mixup: Fairness via interpolation}, in:
  \bibinfo{booktitle}{International Conference on Learning Representations}.
\newblock \URLprefix \url{https://openreview.net/forum?id=DNl5s5BXeBn}.
%Type = Inproceedings
\bibitem[{Chzhen et~al.(2019)Chzhen, Denis, Hebiri, Oneto and
  Pontil}]{chzhen2019leveraging}
\bibinfo{author}{Chzhen, E.}, \bibinfo{author}{Denis, C.},
  \bibinfo{author}{Hebiri, M.}, \bibinfo{author}{Oneto, L.},
  \bibinfo{author}{Pontil, M.}, \bibinfo{year}{2019}.
\newblock \bibinfo{title}{Leveraging labeled and unlabeled data for consistent
  fair binary classification}, in: \bibinfo{booktitle}{Advances in Neural
  Information Processing Systems}, pp. \bibinfo{pages}{12760--12770}.
%Type = Inproceedings
\bibitem[{Corbett-Davies et~al.(2017)Corbett-Davies, Pierson, Feller, Goel and
  Huq}]{corbett2017algorithmic}
\bibinfo{author}{Corbett-Davies, S.}, \bibinfo{author}{Pierson, E.},
  \bibinfo{author}{Feller, A.}, \bibinfo{author}{Goel, S.},
  \bibinfo{author}{Huq, A.}, \bibinfo{year}{2017}.
\newblock \bibinfo{title}{Algorithmic decision making and the cost of
  fairness}, in: \bibinfo{booktitle}{Proceedings of the 23rd acm sigkdd
  international conference on knowledge discovery and data mining}, pp.
  \bibinfo{pages}{797--806}.
%Type = Inproceedings
\bibitem[{Cotter et~al.(2019)Cotter, Jiang and
  Sridharan}]{Cotter2019TwoPlayerGF}
\bibinfo{author}{Cotter, A.}, \bibinfo{author}{Jiang, H.},
  \bibinfo{author}{Sridharan, K.}, \bibinfo{year}{2019}.
\newblock \bibinfo{title}{Two-player games for efficient non-convex constrained
  optimization}, in: \bibinfo{booktitle}{ALT}.
%Type = Inproceedings
\bibitem[{Creager et~al.(2019)Creager, Madras, Jacobsen, Weis, Swersky, Pitassi
  and Zemel}]{creager2019flexibly}
\bibinfo{author}{Creager, E.}, \bibinfo{author}{Madras, D.},
  \bibinfo{author}{Jacobsen, J.H.}, \bibinfo{author}{Weis, M.},
  \bibinfo{author}{Swersky, K.}, \bibinfo{author}{Pitassi, T.},
  \bibinfo{author}{Zemel, R.}, \bibinfo{year}{2019}.
\newblock \bibinfo{title}{Flexibly fair representation learning by
  disentanglement}, in: \bibinfo{booktitle}{International Conference on Machine
  Learning}, \bibinfo{organization}{PMLR}. pp. \bibinfo{pages}{1436--1445}.
%Type = Inproceedings
\bibitem[{Donini et~al.(2018)Donini, Oneto, Ben-David, Shawe-Taylor and
  Pontil}]{donini2018empirical}
\bibinfo{author}{Donini, M.}, \bibinfo{author}{Oneto, L.},
  \bibinfo{author}{Ben-David, S.}, \bibinfo{author}{Shawe-Taylor, J.S.},
  \bibinfo{author}{Pontil, M.}, \bibinfo{year}{2018}.
\newblock \bibinfo{title}{Empirical risk minimization under fairness
  constraints}, in: \bibinfo{booktitle}{Advances in Neural Information
  Processing Systems}, pp. \bibinfo{pages}{2791--2801}.
%Type = Misc
\bibitem[{Dua and Graff(2017)}]{adultdata}
\bibinfo{author}{Dua, D.}, \bibinfo{author}{Graff, C.}, \bibinfo{year}{2017}.
\newblock \bibinfo{title}{{UCI} machine learning repository}.
\newblock \URLprefix \url{http://archive.ics.uci.edu/ml}.
%Type = Inproceedings
\bibitem[{Dwork et~al.(2012)Dwork, Hardt, Pitassi, Reingold and Zemel}]{dwork}
\bibinfo{author}{Dwork, C.}, \bibinfo{author}{Hardt, M.},
  \bibinfo{author}{Pitassi, T.}, \bibinfo{author}{Reingold, O.},
  \bibinfo{author}{Zemel, R.}, \bibinfo{year}{2012}.
\newblock \bibinfo{title}{Fairness through awareness}, in:
  \bibinfo{booktitle}{Proceedings of the 3rd Innovations in Theoretical
  Computer Science Conference}, \bibinfo{publisher}{Association for Computing
  Machinery}, \bibinfo{address}{New York, NY, USA}. p.
  \bibinfo{pages}{214–226}.
\newblock \URLprefix \url{https://doi.org/10.1145/2090236.2090255},
  \DOIprefix\doi{10.1145/2090236.2090255}.
%Type = Inproceedings
\bibitem[{Feldman et~al.(2015)Feldman, Friedler, Moeller, Scheidegger and
  Venkatasubramanian}]{feldman2015certifying}
\bibinfo{author}{Feldman, M.}, \bibinfo{author}{Friedler, S.A.},
  \bibinfo{author}{Moeller, J.}, \bibinfo{author}{Scheidegger, C.},
  \bibinfo{author}{Venkatasubramanian, S.}, \bibinfo{year}{2015}.
\newblock \bibinfo{title}{Certifying and removing disparate impact}, in:
  \bibinfo{booktitle}{proceedings of the 21th ACM SIGKDD international
  conference on knowledge discovery and data mining}, pp.
  \bibinfo{pages}{259--268}.
%Type = Inproceedings
\bibitem[{Fish et~al.(2016)Fish, Kun and Lelkes}]{fish2016confidence}
\bibinfo{author}{Fish, B.}, \bibinfo{author}{Kun, J.}, \bibinfo{author}{Lelkes,
  {\'A}.D.}, \bibinfo{year}{2016}.
\newblock \bibinfo{title}{A confidence-based approach for balancing fairness
  and accuracy}, in: \bibinfo{booktitle}{Proceedings of the 2016 SIAM
  International Conference on Data Mining}, \bibinfo{organization}{SIAM}. pp.
  \bibinfo{pages}{144--152}.
%Type = Inproceedings
\bibitem[{van~de Geer(2000)}]{Geer2000EmpiricalPI}
\bibinfo{author}{van~de Geer, S.A.}, \bibinfo{year}{2000}.
\newblock \bibinfo{title}{Empirical processes in m-estimation}.
%Type = Inproceedings
\bibitem[{Goh et~al.(2016)Goh, Cotter, Gupta and
  Friedlander}]{goh2016satisfying}
\bibinfo{author}{Goh, G.}, \bibinfo{author}{Cotter, A.},
  \bibinfo{author}{Gupta, M.}, \bibinfo{author}{Friedlander, M.P.},
  \bibinfo{year}{2016}.
\newblock \bibinfo{title}{Satisfying real-world goals with dataset
  constraints}, in: \bibinfo{booktitle}{Advances in Neural Information
  Processing Systems}, pp. \bibinfo{pages}{2415--2423}.
%Type = Inproceedings
\bibitem[{Hardt et~al.(2016)Hardt, Price and Srebro}]{hardt2016equality}
\bibinfo{author}{Hardt, M.}, \bibinfo{author}{Price, E.},
  \bibinfo{author}{Srebro, N.}, \bibinfo{year}{2016}.
\newblock \bibinfo{title}{Equality of opportunity in supervised learning}, in:
  \bibinfo{booktitle}{Advances in neural information processing systems}, pp.
  \bibinfo{pages}{3315--3323}.
%Type = Inproceedings
\bibitem[{Jiang et~al.(2020)Jiang, Pacchiano, Stepleton, Jiang and
  Chiappa}]{jiang2020wasserstein}
\bibinfo{author}{Jiang, R.}, \bibinfo{author}{Pacchiano, A.},
  \bibinfo{author}{Stepleton, T.}, \bibinfo{author}{Jiang, H.},
  \bibinfo{author}{Chiappa, S.}, \bibinfo{year}{2020}.
\newblock \bibinfo{title}{Wasserstein fair classification}, in:
  \bibinfo{booktitle}{Uncertainty in Artificial Intelligence},
  \bibinfo{organization}{PMLR}. pp. \bibinfo{pages}{862--872}.
%Type = Inproceedings
\bibitem[{Kamiran et~al.(2012)Kamiran, Karim and Zhang}]{kamiran2012decision}
\bibinfo{author}{Kamiran, F.}, \bibinfo{author}{Karim, A.},
  \bibinfo{author}{Zhang, X.}, \bibinfo{year}{2012}.
\newblock \bibinfo{title}{Decision theory for discrimination-aware
  classification}, in: \bibinfo{booktitle}{2012 IEEE 12th International
  Conference on Data Mining}, \bibinfo{organization}{IEEE}. pp.
  \bibinfo{pages}{924--929}.
%Type = Inproceedings
\bibitem[{Kamishima et~al.(2012)Kamishima, Akaho, Asoh and
  Sakuma}]{kamishima2012fairness}
\bibinfo{author}{Kamishima, T.}, \bibinfo{author}{Akaho, S.},
  \bibinfo{author}{Asoh, H.}, \bibinfo{author}{Sakuma, J.},
  \bibinfo{year}{2012}.
\newblock \bibinfo{title}{Fairness-aware classifier with prejudice remover
  regularizer}, in: \bibinfo{booktitle}{Joint European Conference on Machine
  Learning and Knowledge Discovery in Databases},
  \bibinfo{organization}{Springer}. pp. \bibinfo{pages}{35--50}.
%Type = Inproceedings
\bibitem[{Kingma and Ba(2015)}]{adam}
\bibinfo{author}{Kingma, D.P.}, \bibinfo{author}{Ba, J.}, \bibinfo{year}{2015}.
\newblock \bibinfo{title}{Adam: A method for stochastic optimization}, in:
  \bibinfo{editor}{Bengio, Y.}, \bibinfo{editor}{LeCun, Y.} (Eds.),
  \bibinfo{booktitle}{3rd International Conference on Learning Representations,
  {ICLR} 2015, San Diego, CA, USA, May 7-9, 2015, Conference Track
  Proceedings}.
\newblock \URLprefix \url{http://arxiv.org/abs/1412.6980}.
%Type = Inproceedings
\bibitem[{Kleinberg et~al.(2018)Kleinberg, Ludwig, Mullainathan and
  Rambachan}]{kleinberg2018algorithmic}
\bibinfo{author}{Kleinberg, J.}, \bibinfo{author}{Ludwig, J.},
  \bibinfo{author}{Mullainathan, S.}, \bibinfo{author}{Rambachan, A.},
  \bibinfo{year}{2018}.
\newblock \bibinfo{title}{Algorithmic fairness}, in: \bibinfo{booktitle}{Aea
  papers and proceedings}, pp. \bibinfo{pages}{22--27}.
%Type = Misc
\bibitem[{Lichman(2013)}]{uci}
\bibinfo{author}{Lichman, M.}, \bibinfo{year}{2013}.
\newblock \bibinfo{title}{{UCI} machine learning repository}.
\newblock \URLprefix \url{http://archive.ics.uci.edu/ml}.
%Type = Inproceedings
\bibitem[{Lohaus et~al.(2020)Lohaus, Perrot and Luxburg}]{pmlr-v119-lohaus20a}
\bibinfo{author}{Lohaus, M.}, \bibinfo{author}{Perrot, M.},
  \bibinfo{author}{Luxburg, U.V.}, \bibinfo{year}{2020}.
\newblock \bibinfo{title}{Too relaxed to be fair}, in: \bibinfo{editor}{III,
  H.D.}, \bibinfo{editor}{Singh, A.} (Eds.), \bibinfo{booktitle}{Proceedings of
  the 37th International Conference on Machine Learning},
  \bibinfo{publisher}{PMLR}. pp. \bibinfo{pages}{6360--6369}.
\newblock \URLprefix \url{https://proceedings.mlr.press/v119/lohaus20a.html}.
%Type = Inproceedings
\bibitem[{Madras et~al.(2018)Madras, Creager, Pitassi and
  Zemel}]{Madras2018LearningAF}
\bibinfo{author}{Madras, D.}, \bibinfo{author}{Creager, E.},
  \bibinfo{author}{Pitassi, T.}, \bibinfo{author}{Zemel, R.S.},
  \bibinfo{year}{2018}.
\newblock \bibinfo{title}{Learning adversarially fair and transferable
  representations}, in: \bibinfo{booktitle}{ICML}.
%Type = Article
\bibitem[{Mahendhiran and Subramanian(2018)}]{article-unknow}
\bibinfo{author}{Mahendhiran, P.}, \bibinfo{author}{Subramanian, K.},
  \bibinfo{year}{2018}.
\newblock \bibinfo{title}{Deep learning techniques for polarity classification
  in multimodal sentiment analysis}.
\newblock \bibinfo{journal}{International Journal of Information Technology and
  Decision Making} \bibinfo{volume}{17}.
\newblock \DOIprefix\doi{10.1142/S0219622018500128}.
%Type = Misc
\bibitem[{Maity et~al.(2021)Maity, Xue, Yurochkin and
  Sun}]{https://doi.org/10.48550/arxiv.2103.16714}
\bibinfo{author}{Maity, S.}, \bibinfo{author}{Xue, S.},
  \bibinfo{author}{Yurochkin, M.}, \bibinfo{author}{Sun, Y.},
  \bibinfo{year}{2021}.
\newblock \bibinfo{title}{Statistical inference for individual fairness}.
\newblock \URLprefix \url{https://arxiv.org/abs/2103.16714},
  \DOIprefix\doi{10.48550/ARXIV.2103.16714}.
%Type = Article
\bibitem[{Mehrabi et~al.(2019)Mehrabi, Morstatter, Saxena, Lerman and
  Galstyan}]{mehrabi2019survey}
\bibinfo{author}{Mehrabi, N.}, \bibinfo{author}{Morstatter, F.},
  \bibinfo{author}{Saxena, N.}, \bibinfo{author}{Lerman, K.},
  \bibinfo{author}{Galstyan, A.}, \bibinfo{year}{2019}.
\newblock \bibinfo{title}{A survey on bias and fairness in machine learning}.
\newblock \bibinfo{journal}{arXiv preprint arXiv:1908.09635} .
%Type = Misc
\bibitem[{Miyato et~al.(2018)Miyato, ichi Maeda, Koyama and Ishii}]{vat}
\bibinfo{author}{Miyato, T.}, \bibinfo{author}{ichi Maeda, S.},
  \bibinfo{author}{Koyama, M.}, \bibinfo{author}{Ishii, S.},
  \bibinfo{year}{2018}.
\newblock \bibinfo{title}{Virtual adversarial training: A regularization method
  for supervised and semi-supervised learning}.
\newblock \href{http://arxiv.org/abs/1704.03976}{\tt arXiv:1704.03976}.
%Type = Inproceedings
\bibitem[{Pleiss et~al.(2017)Pleiss, Raghavan, Wu, Kleinberg and
  Weinberger}]{pleiss2017fairness}
\bibinfo{author}{Pleiss, G.}, \bibinfo{author}{Raghavan, M.},
  \bibinfo{author}{Wu, F.}, \bibinfo{author}{Kleinberg, J.},
  \bibinfo{author}{Weinberger, K.Q.}, \bibinfo{year}{2017}.
\newblock \bibinfo{title}{On fairness and calibration}, in:
  \bibinfo{booktitle}{Advances in Neural Information Processing Systems}, pp.
  \bibinfo{pages}{5680--5689}.
%Type = Article
\bibitem[{Schmidt-Hieber et~al.(2020)}]{schmidt2020nonparametric}
\bibinfo{author}{Schmidt-Hieber, J.}, et~al., \bibinfo{year}{2020}.
\newblock \bibinfo{title}{Nonparametric regression using deep neural networks
  with relu activation function}.
\newblock \bibinfo{journal}{Annals of Statistics} \bibinfo{volume}{48},
  \bibinfo{pages}{1875--1897}.
%Type = Book
\bibitem[{Shalev-Shwartz and Ben-David(2014)}]{shalev2014understanding}
\bibinfo{author}{Shalev-Shwartz, S.}, \bibinfo{author}{Ben-David, S.},
  \bibinfo{year}{2014}.
\newblock \bibinfo{title}{Understanding machine learning: From theory to
  algorithms}.
\newblock \bibinfo{publisher}{Cambridge university press}.
%Type = Article
\bibitem[{Shen et~al.(2003)Shen, Tseng, Zhang and Wong}]{shen2003psi}
\bibinfo{author}{Shen, X.}, \bibinfo{author}{Tseng, G.C.},
  \bibinfo{author}{Zhang, X.}, \bibinfo{author}{Wong, W.H.},
  \bibinfo{year}{2003}.
\newblock \bibinfo{title}{On $\psi$-learning}.
\newblock \bibinfo{journal}{Journal of the American Statistical Association}
  \bibinfo{volume}{98}, \bibinfo{pages}{724--734}.
%Type = Inproceedings
\bibitem[{Suzuki(2019)}]{suzuki2018adaptivity}
\bibinfo{author}{Suzuki, T.}, \bibinfo{year}{2019}.
\newblock \bibinfo{title}{Adaptivity of deep re{LU} network for learning in
  besov and mixed smooth besov spaces: optimal rate and curse of
  dimensionality}, in: \bibinfo{booktitle}{International Conference on Learning
  Representations}.
\newblock \URLprefix \url{https://openreview.net/forum?id=H1ebTsActm}.
%Type = Article
\bibitem[{Vogel et~al.(2020)Vogel, Bellet and
  Cl{\'e}men{\c{c}}on}]{vogel2020learning}
\bibinfo{author}{Vogel, R.}, \bibinfo{author}{Bellet, A.},
  \bibinfo{author}{Cl{\'e}men{\c{c}}on, S.}, \bibinfo{year}{2020}.
\newblock \bibinfo{title}{Learning {F}air {S}coring {F}unctions: {F}airness
  {D}efinitions, {A}lgorithms and {G}eneralization {B}ounds for {B}ipartite
  {R}anking}.
\newblock \bibinfo{journal}{arXiv preprint arXiv:2002.08159} .
%Type = Article
\bibitem[{Webster et~al.(2018)Webster, Recasens, Axelrod and
  Baldridge}]{webster2018mind}
\bibinfo{author}{Webster, K.}, \bibinfo{author}{Recasens, M.},
  \bibinfo{author}{Axelrod, V.}, \bibinfo{author}{Baldridge, J.},
  \bibinfo{year}{2018}.
\newblock \bibinfo{title}{Mind the gap: A balanced corpus of gendered ambiguous
  pronouns}.
\newblock \bibinfo{journal}{Transactions of the Association for Computational
  Linguistics} \bibinfo{volume}{6}, \bibinfo{pages}{605--617}.
%Type = Inproceedings
\bibitem[{Wei et~al.(2020)Wei, Ramamurthy and Calmon}]{wei20a}
\bibinfo{author}{Wei, D.}, \bibinfo{author}{Ramamurthy, K.N.},
  \bibinfo{author}{Calmon, F.}, \bibinfo{year}{2020}.
\newblock \bibinfo{title}{Optimized score transformation for fair
  classification}, \bibinfo{publisher}{PMLR}, \bibinfo{address}{Online}. pp.
  \bibinfo{pages}{1673--1683}.
\newblock \URLprefix \url{http://proceedings.mlr.press/v108/wei20a.html}.
%Type = Misc
\bibitem[{Wolf(2018)}]{wolf2018mathematical}
\bibinfo{author}{Wolf, M.M.}, \bibinfo{year}{2018}.
\newblock \bibinfo{title}{Mathematical foundations of supervised learning}.
%Type = Inproceedings
\bibitem[{Woodworth et~al.(2017)Woodworth, Gunasekar, Ohannessian and
  Srebro}]{pmlr-v65-woodworth17a}
\bibinfo{author}{Woodworth, B.}, \bibinfo{author}{Gunasekar, S.},
  \bibinfo{author}{Ohannessian, M.I.}, \bibinfo{author}{Srebro, N.},
  \bibinfo{year}{2017}.
\newblock \bibinfo{title}{Learning non-discriminatory predictors}, in:
  \bibinfo{editor}{Kale, S.}, \bibinfo{editor}{Shamir, O.} (Eds.),
  \bibinfo{booktitle}{Proceedings of the 2017 Conference on Learning Theory},
  \bibinfo{publisher}{PMLR}. pp. \bibinfo{pages}{1920--1953}.
\newblock \URLprefix \url{http://proceedings.mlr.press/v65/woodworth17a.html}.
%Type = Inproceedings
\bibitem[{Wu et~al.(2019)Wu, Zhang and Wu}]{onconvexand}
\bibinfo{author}{Wu, Y.}, \bibinfo{author}{Zhang, L.}, \bibinfo{author}{Wu,
  X.}, \bibinfo{year}{2019}.
\newblock \bibinfo{title}{On convexity and bounds of fairness-aware
  classification}, in: \bibinfo{booktitle}{The World Wide Web Conference},
  \bibinfo{publisher}{Association for Computing Machinery},
  \bibinfo{address}{New York, NY, USA}. p. \bibinfo{pages}{3356–3362}.
\newblock \URLprefix \url{https://doi.org/10.1145/3308558.3313723},
  \DOIprefix\doi{10.1145/3308558.3313723}.
%Type = Inproceedings
\bibitem[{Xu et~al.(2018)Xu, Yuan, Zhang and Wu}]{xu2018fairgan}
\bibinfo{author}{Xu, D.}, \bibinfo{author}{Yuan, S.}, \bibinfo{author}{Zhang,
  L.}, \bibinfo{author}{Wu, X.}, \bibinfo{year}{2018}.
\newblock \bibinfo{title}{Fairgan: Fairness-aware generative adversarial
  networks}, in: \bibinfo{booktitle}{2018 IEEE International Conference on Big
  Data (Big Data)}, \bibinfo{organization}{IEEE}. pp.
  \bibinfo{pages}{570--575}.
%Type = Inproceedings
\bibitem[{Yona and Rothblum(2018)}]{pacf}
\bibinfo{author}{Yona, G.}, \bibinfo{author}{Rothblum, G.},
  \bibinfo{year}{2018}.
\newblock \bibinfo{title}{Probably approximately metric-fair learning}, in:
  \bibinfo{editor}{Dy, J.}, \bibinfo{editor}{Krause, A.} (Eds.),
  \bibinfo{booktitle}{Proceedings of the 35th International Conference on
  Machine Learning}, \bibinfo{publisher}{PMLR},
  \bibinfo{address}{Stockholmsmässan, Stockholm Sweden}. pp.
  \bibinfo{pages}{5680--5688}.
%Type = Inproceedings
\bibitem[{Yuille and Rangarajan(2002)}]{cccp}
\bibinfo{author}{Yuille, A.L.}, \bibinfo{author}{Rangarajan, A.},
  \bibinfo{year}{2002}.
\newblock \bibinfo{title}{The concave-convex procedure (cccp)}, in:
  \bibinfo{editor}{Dietterich, T.}, \bibinfo{editor}{Becker, S.},
  \bibinfo{editor}{Ghahramani, Z.} (Eds.), \bibinfo{booktitle}{Advances in
  Neural Information Processing Systems}, \bibinfo{publisher}{MIT Press}. pp.
  \bibinfo{pages}{1033--1040}.
%Type = Inproceedings
\bibitem[{Yurochkin et~al.(2020)Yurochkin, Bower and
  Sun}]{Yurochkin2020Training}
\bibinfo{author}{Yurochkin, M.}, \bibinfo{author}{Bower, A.},
  \bibinfo{author}{Sun, Y.}, \bibinfo{year}{2020}.
\newblock \bibinfo{title}{Training individually fair ml models with sensitive
  subspace robustness}, in: \bibinfo{booktitle}{International Conference on
  Learning Representations}.
\newblock \URLprefix \url{https://openreview.net/forum?id=B1gdkxHFDH}.
%Type = Misc
\bibitem[{Yurochkin and Sun(2020)}]{sensei}
\bibinfo{author}{Yurochkin, M.}, \bibinfo{author}{Sun, Y.},
  \bibinfo{year}{2020}.
\newblock \bibinfo{title}{Sensei: Sensitive set invariance for enforcing
  individual fairness}.
\newblock \href{http://arxiv.org/abs/2006.14168}{\tt arXiv:2006.14168}.
%Type = Article
\bibitem[{Zafar et~al.(2019)Zafar, Valera, Gomez-Rodriguez and
  Gummadi}]{zafar2019fairness}
\bibinfo{author}{Zafar, M.B.}, \bibinfo{author}{Valera, I.},
  \bibinfo{author}{Gomez-Rodriguez, M.}, \bibinfo{author}{Gummadi, K.P.},
  \bibinfo{year}{2019}.
\newblock \bibinfo{title}{Fairness {C}onstraints: A {F}lexible {A}pproach for
  {F}air {C}lassification.}
\newblock \bibinfo{journal}{J. Mach. Learn. Res.} \bibinfo{volume}{20},
  \bibinfo{pages}{1--42}.
%Type = Inproceedings
\bibitem[{Zafar et~al.(2017)Zafar, Valera, Rogriguez and
  Gummadi}]{zafar2017fairness}
\bibinfo{author}{Zafar, M.B.}, \bibinfo{author}{Valera, I.},
  \bibinfo{author}{Rogriguez, M.G.}, \bibinfo{author}{Gummadi, K.P.},
  \bibinfo{year}{2017}.
\newblock \bibinfo{title}{Fairness constraints: Mechanisms for fair
  classification}, in: \bibinfo{booktitle}{Artificial Intelligence and
  Statistics}, pp. \bibinfo{pages}{962--970}.
%Type = Inproceedings
\bibitem[{Zemel et~al.(2013)Zemel, Wu, Swersky, Pitassi and
  Dwork}]{zemel2013learning}
\bibinfo{author}{Zemel, R.}, \bibinfo{author}{Wu, Y.},
  \bibinfo{author}{Swersky, K.}, \bibinfo{author}{Pitassi, T.},
  \bibinfo{author}{Dwork, C.}, \bibinfo{year}{2013}.
\newblock \bibinfo{title}{Learning fair representations}, in:
  \bibinfo{booktitle}{International Conference on Machine Learning}, pp.
  \bibinfo{pages}{325--333}.
%Type = Article
\bibitem[{Zhang(2004)}]{zhang2004statistical}
\bibinfo{author}{Zhang, T.}, \bibinfo{year}{2004}.
\newblock \bibinfo{title}{Statistical behavior and consistency of
  classification methods based on convex risk minimization}.
\newblock \bibinfo{journal}{The Annals of Statistics} \bibinfo{volume}{32},
  \bibinfo{pages}{56--85}.

\end{thebibliography}
	
	\bio{}
	\endbio
	
	\endbio
	
\end{document}